\documentclass[runningheads]{llncs}
\usepackage{graphicx}
\usepackage{microtype}
\usepackage{subfigure}
\usepackage{hyperref}
\usepackage[bibliography=separate]{apxproof}
\renewcommand{\appendixsectionformat}[2]{Supplementary Material for Section #1 (#2)}
\usepackage{mathdefs_deb}
\allowdisplaybreaks
\begin{document}
\title{BelMan: An Information-Geometric Approach to Stochastic Bandits}
%
%
\author{Debabrota Basu\inst{1} \and
Pierre Senellart\inst{2,3} \and
St{\'e}phane Bressan\inst{4}}
\authorrunning{Basu et al.}
%
\institute{Data Science and AI Division, Chalmers University of Technology, G\"{o}teborg, Sweden \and DI ENS, ENS, CNRS, PSL University, Paris, France \and Inria, Paris, France\and School of Computing, National University of Singapore, Singapore}
\maketitle              
\begin{abstract}
We propose a Bayesian information-geometric approach to the exploration--exploitation trade-off in stochastic multi-armed bandits. The uncertainty on reward generation and belief is represented using the manifold of joint distributions of rewards and beliefs. Accumulated information is summarised by the barycentre of joint distributions, the \emph{pseudobelief-reward}. While the pseudobelief-reward facilitates information accumulation through exploration, another mechanism is needed to increase exploitation by gradually focusing on higher rewards, the \emph{pseudobelief-focal-reward}. Our resulting algorithm, BelMan, alternates between projection of the pseudobelief-focal-reward onto belief-reward distributions to choose the arm to play, and projection of the updated belief-reward distributions onto the pseudobelief-focal-reward. We theoretically prove BelMan to be asymptotically optimal and to incur a sublinear regret growth. We instantiate BelMan to stochastic bandits with Bernoulli and exponential rewards, and to a real-life application of scheduling queueing bandits. Comparative evaluation with the state of the art shows that BelMan is not only competitive for Bernoulli bandits but in many cases also outperforms other approaches for exponential and queueing bandits.
\end{abstract}


%
\begin{toappendix}
	\centerline{\textbf{\large Supplementary Material: BelMan}}
\bigskip
		We provide the following supplementary material:
		\begin{itemize}
                  \item in Section~\ref{app:related}, an extended discussion of the related work and of the setting
			of bandits, beyond what could
			fit in the main paper;
                      \item in Section~\ref{app:proofs}, proofs and
                        technical details that complement the methodology
			section (Section~\ref{sec:method});
                      \item in Section~\ref{app:experiments}, additional experiments in the exploration--exploitation setup.
		\end{itemize}
\end{toappendix}
\section{Introduction}\label{sec:intro}
The \emph{multi-armed bandit} problem~\cite{robbins1952} is a sequential decision-making problem~\cite{degroot2005optimal} in which a gambler
plays a set of arms to obtain a sequence of rewards. In the
\emph{stochastic bandit} problem~\cite{bubeck2012book}, the rewards are
obtained from reward distributions on arms. These reward distributions
belong to the same family of distributions but vary in the parameters. These parameters are unknown to the gambler.
In the classical setting, the gambler devises a strategy, choosing a sequence of arm draws, that maximises the \emph{expected cumulative
reward}~\cite{robbins1952}. 
In an equivalent formulation, the gambler devises a strategy that minimises the \emph{expected cumulative
regret}~\cite{lairobbins1985}, that is the expected cumulative deficit of reward caused by the gambler not always playing the optimal arm. 
In order to achieve this goal, the gambler must simultaneously learn the parameters of the reward distributions of arms. 
Thus, solving the stochastic bandit problem consists in devising
strategies that combine both the accumulation of information to reduce
the uncertainty of decision making, \emph{exploration}, and the
accumulation of rewards, \emph{exploitation}~\cite{macready1998exex}. We
refer to the stochastic bandit problem as the
\emph{exploration--exploitation bandit} problem to highlight this trade-off. 
If a strategy relies on independent phases of exploration and exploitation, it necessarily yields a suboptimal regret bound~\cite{garivier2016etc}. Gambler has to adaptively balance and intertwine exploration and exploitation~\cite{auer2002finite}. 

In a variant of the stochastic bandit problem, called the \emph{pure
exploration bandit} problem~\cite{bubeck2009pure}, the goal of the
gambler is solely to accumulate information about the arms. In another
variant of the stochastic bandit problem, the gambler interacts with the
bandit in two consecutive phases of pure exploration and exploration--exploitation. The authors of \cite{putta2017twophaseepisodic} named this variant the \emph{two-phase reinforcement learning} problem.

Although frequentist algorithms with optimism in the face of uncertainty such as UCB~\cite{auer2002finite} and KL-UCB~\cite{garivier2011klucb} work considerably well for the exploration--exploitation bandit problem, their frequentist nature prevents effective assimilation of a priori knowledge about the reward distributions of the arms~\cite{kawale2015efficient}.
Bayesian algorithms for the exploration--exploitation problem, such as
Thompson sampling~\cite{thompson1933} and
Bayes-UCB~\cite{kaufmann2012bayesucb}, leverage a prior distribution that
summarises a priori knowledge.
However, as argued in~\cite{kaufmann2013information}, there is a need for Bayesian algorithms that also cater for pure exploration. Neither Thompson sampling nor Bayes-UCB are able to do so.

\textbf{Our contribution.} We propose a unified Bayesian approach to
address the exploration--exploitation, pure exploration, and two-phase reinforcement learning problems.
We address these problems from the perspective of information representation, accumulation, and balanced induction of bias. Here, the uncertainty is two fold. Sampling reward from the reward distributions is inherently stochastic. The other layer is due to the incomplete information about the true paramaters of the reward distributions. Following Bayesian algorithms~\cite{thompson1933}, we maintain a parameterised \emph{belief} distribution for each arm representing
the uncertainty on the parameter of its reward distribution. Extending this representation, we use a joint distribution to express the two-fold uncertainty induced by both the belief and the reward distributions of each arm. We refer to these joint distributions as the \emph{belief-reward distributions} of the arms.
We set the learning problem in the statistical manifold~\cite{amari2007infogeo} of the belief-reward distributions, which we call the \emph{belief-reward manifold}. 
The belief-reward manifold provides a representation for controlling pure exploration and exploration--exploitation, and to design a unifying algorithmic framework. 

The authors of \cite{bubeck2009pure} proved that, for Bernoulli bandits,
if an exploration--exploitation algorithm achieves an upper-bounded
regret, it cannot reduce the expected simple regret by more than a fixed
lower bound. This drives us to first devise a pure exploration algorithm,
which requires a collective representation of the accumulated knowledge about the arm.
From an information-geometric point of view~\cite{barbaresco2013information,agueh2011barycenters}, the barycentre of the belief-reward distributions in the belief-reward manifolds serves as a succinct summary. We refer to this barycentre as the \emph{pseudobelief-reward}. 
We prove the pseudobelief-reward to be a unique representation in the manifold. 
Though pseudobelief-reward facilitates the accumulation of knowledge, it is essential for the exploration–exploitation bandit problem to also incorporate a mechanism that gradually concentrates on  higher rewards~\cite{macready1998exex}.
We introduce a distribution that induces such an increasing exploitative bias. We refer to this distribution as the \emph{focal distribution}. We incorporate it into the definition of the pseudobelief-reward distribution to construct the \emph{pseudobelief-focal-reward distribution}. This pushes the summarised representation towards the arms having higher expected rewards. We implement the focal distribution using an exponential function of the form $\exp(X/\tau(t))$,
where $X$ is the reward, and a parameter $\tau(t)$ dependent on time $t$ and named as \emph{exposure}. Exposure controls the exploration--exploitation trade-off.

In Section~\ref{sec:method}, we apply these information-geometric constructions to develop the BelMan
algorithm. BelMan projects the pseudobelief-focal-reward onto belief-rewards to select an arm. As it is played and a reward is collected, BelMan updates the belief-reward distribution of the corresponding arm by projecting of the updated belief-reward distributions onto the pseudobelief-focal-reward. 
Information geometrically these two projections are studied as information (I-) and reverse information (rI-) projections~\cite{csiszar1984iproj}, respectively.
BelMan alternates I- and rI-projections between belief-reward distributions of the arms and the pseudobelief-focal-reward distribution for arm selection and information accumulation.
We prove the law of convergence of the pseudobelief-focal-reward distribution for BelMan, and that BelMan asymptotically converges to the choice of the optimal arm.
BelMan can be tuned, using the exposure, to support a continuum from pure
exploration to exploration--exploitation, as well as two-phase reinforcement learning.

We instantiate BelMan for distributions of the exponential family~\cite{brown1986expfamily}. These distributions lead to analytical forms that allows derivation of well-defined and unique I- and rI-projections as well as to devise an effective and fast computation.
In Section~\ref{sec:parametric}, we empirically evaluate the performance of BelMan on different sets of arms and parameters for Bernoulli and exponential distributions, thus showing its applicability to both discrete and continuous rewards.
Experimental results  validate that BelMan asymptotically achieves
logarithmic regret. We compare BelMan with state-of-the-art algorithms:
UCB~\cite{auer2002finite}, KL-UCB, KL-UCB-Exp~\cite{garivier2011klucb},
Bayes-UCB~\cite{kaufmann2012bayesucb}, Thompson
sampling~\cite{thompson1933}, and Gittins index~\cite{gittins1979}, in
these different settings.  Results demonstrate that BelMan is not only
competitive but also outperforms existing algorithms for challenging
setups such as those involving many arms and continuous rewards. For the
two-phase reinforcement learning, results show that BelMan spontaneously
adapts to the explored information, improving the efficiency.

We also instantiate BelMan to the application of queueing bandits~\cite{krishnasamy2016regret}. 
Queueing bandits represent the problem of scheduling jobs in a
multi-server queueing system with unknown service rates. The goal of the corresponding scheduling algorithm is to minimise the number of jobs in hold while also learning the service rates.
A comparative performance evaluation for queueing systems with Bernoulli service rates show that BelMan performs significantly better than the existing algorithms, such as Q-UCB, Q-ThS, and Thompson sampling.


\begin{toappendix}
	\section{Extended Discussion of Related Work}
        \label{app:related}
	\paragraph{Exploration--exploitation bandit problem.} In the exploration--exploitation bandits, the agent searches for a policy that maximises the expected value of \textit{cumulative reward} $S_T \triangleq \mathbb{E} \left[\sum_{t=1}^T X_{a_t}\right]$ as $T \rightarrow \infty$.
A policy is \textit{asymptotically consistent}~\cite{robbins1952} if it asymptotically
tends to choose the arm with maximum expected reward $\mu^* \triangleq \max_{1\leq a \leq K} \mu(\theta_a)$, i.e.,
\begin{equation}\label{eqn:asymp_conv}
\lim_{T \rightarrow \infty} \frac{1}{T} S_T = \mu^*.
\end{equation}
The \emph{cumulative regret} $R_T$~\cite{lairobbins1985} is the amount of extra reward the gambler can obtain if she knows the optimal arm $a^*$ and always plays it instead of the present sequence:
\begin{align*}
R_T &\triangleq T \mathbb{E}\left[X_{a^*}\right] - \mathbb{E}\left[\sum_{t=1}^T \left(X_{a_t}\right)\right]\\ &= T \mu^* - \sum_{a=1}^K \mathbb{E}\left[\sum_{t=1}^T \left(X_{a_t} \times \bm{1}(a_t=a)\right)\right]\\ &= \sum_{a=1}^K \left[ \mu^* - \mu^a  \right]\mathbb{E}[t_{T}^{a}], 
\end{align*}
where $t_{T}^{a}$ is the number of times arm $a$ is pulled till the $T$th iteration.
\cite{lairobbins1985} proved that for all algorithms satisfying $R_T = o(T^c)$ for a non-negative $c$, the cumulative regret increases asymptotically in $\Omega(\log T)$.
Such algorithms are called \textit{asymptotically efficient}.
The Lai--Robbins bound can be mathematically formulated as
\begin{equation}\label{eqn:lr_bound}
    \mathop{\lim\inf}\limits_{T \rightarrow \infty} \frac{R_T}{\log T} \geq \frac{\sum\limits_{a:\mu^*(\bm{\theta}) > \mu({\theta}_a)} \left[ \mu^*(\bm{\theta}) - \mu(\theta_a) \right]}{\inf_{a} D_{\mathrm{KL}}(f_{\theta_a}(x)\parallel f_{\theta^*}(x)) },
  \end{equation}
where $f_{\theta^*}(x)$ is the reward distribution of the optimal arm.
This states that the best we can achieve is a logarithmic growth of cumulative regret.
It also implies that this optimality is harder to achieve as the minimal KL-divergence between the optimal arm and any other arm decreases.
This is intuitive because in such scenario the agent has to explore these two arms more to distinguish between them and to choose the optimal arm.
\cite{lairobbins1985} also showed that for specific reward distributions, the expected number of draws of any suboptimal arm $a$ should satisfy
  \begin{equation}\label{eqn:subopt_bound}
t_{T}^{a} \leq \left( \frac{1}{\inf_{a} D_{\mathrm{KL}}(f_{\theta_a}(x)\parallel f_{\theta^*}(x)) } + o(1) \right) \log T.
\end{equation} 
Equation~\eqref{eqn:lr_bound} and~\eqref{eqn:subopt_bound} together claim that the best achievable number of draws of suboptimal arms is $\Theta(\log T)$.
Based on this bound, \cite{auer2002finite} extensively studied the upper confidence bound (UCB) family of algorithms.
These algorithms operate on the philosophy of optimism in face of uncertainty.
They compute the upper confidence bounds of each of the arm's distributions in a frequentist way and choose the one with the maximum upper confidence bound optimistically expecting that one to be the arm with maximum expected reward.
Later on, this family of algorithms was analysed and improved to propose algorithms such as KL-UCB~\cite{garivier2011klucb} and DMED~\cite{honda2011asymptotically}.

Frequentist approaches implicitly assume a `true' parametrization of reward distributions $\left( \theta_1^{true}, \ldots, \theta_K^{true}\right)$.
In contrast, Bayesians model the uncertainty on the parameter using another probability distribution $B\left( \theta_1, \ldots, \theta_K\right)$ \cite{degroot2005optimal,scott2010modernmab} which is referred to as the
\textit{belief distribution}.
Bayesian algorithms begin with a prior $B_0 \left( \theta_1, \ldots, \theta_K\right)$ over the parameters and eventually try to find out a posterior distribution such that the Bayesian sum of rewards $\int S_T {\mathrm{d}B\left( \theta_1, \ldots, \theta_K\right)}$ is maximised, or equivalently the Bayesian risk $\int R_T{\mathrm{d}B\left( \theta_1, \ldots, \theta_K\right)}$ is minimised.

Another variant of the Bayesian formulation was introduced by~\cite{bellman1956} with a discounted reward setting.
Unlike $S_T$, the discounted sum of rewards $D_\gamma \triangleq \sum_{t=0}^\infty \left[ \gamma^t x_{t+1} \right]$ is calculated over infinite horizon. Here, $\gamma \in [0, 1)$ ensures convergence of the sequential sum of rewards for infinite horizon.
Intuitively, the discounted sum implies the effect of an action decay with each time step by the discount factor $\gamma$.
This setting assumes $K$ independent priors on each of the arms and also models the process of choosing the next arm as a Markov process.
Thus, the bandit problem is reformulated as maximising
  \[
\int \ldots \int \mathbb{E}_{\bm{\theta}}[D_\gamma] {\mathrm{d}b^1(\theta_1)} \ldots {\mathrm{d}b^K(\theta_K)}
  \]
where, $b^a$ is the independent prior distribution on the parameter $\theta_a$ for $a = 1, \ldots, K$.
\cite{gittins1979} showed the agent can have an optimally indexed policy by sampling from the arm with largest Gittins index
  \[
G^a (s^a) \triangleq \sup_{\tau > 0} \frac{\mathbb{E} \left[\sum\limits_{t = 0}^{\tau} \gamma^t x^a (S^a_t) \mid S^a_0 = s^a  \right]}{\mathbb{E} \left[ \sum\limits_{t=0}^{\tau - 1} \gamma^t \mid S^a_0 = s^a\right]}
  \]
where $s^a$ is the state of arm $a$ and $\tau$ is referred to as the stopping time i.e, the first time when the index is no greater than its initial value.
Though Gittins index \cite{gittins1979} is proven to be optimal for discounted Bayesian bandits with Bernoulli rewards, explicit computation of the indices is not always tractable and does not provide clear insights into what they look like and how they change as sampling proceeds~\cite{nino2011computing}.

Thus, researchers developed approximation algorithms~\cite{lai1988asymptotic} and sequential sampling schemes like Thompson sampling~\cite{thompson1933}.
At any iteration, the latter samples $K$ parameter values from the belief
distributions and chooses the arm that has maximum expected reward for them.
\cite{kaufmann2012bayesucb} also proposed a Bayesian analogue of the UCB algorithm.
Unlike the original, it uses belief distributions to keep
track of arm uncertainty and update them using Bayes' theorem,
computes UCBs for each arm using the
belief distributions, and chooses the arm accordingly.

\paragraph{Pure exploration bandit problem.} In this variant of the bandit problem, the agent aims to gain more information about the arms.
\cite{bubeck2009pure} formulated this notion of gaining information as minimisation of the simple regret rather than cumulative regret.
\textit{Simple regret} $r_t (\bm{\theta})$ at time $t$ is the expected difference between the maximum achievable reward~$X_{a^*}$ and the sampled reward $X_{a_t}$.
Unlike cumulative regret, minimising simple regret depends only on exploration and the number of available rounds to do so.
\cite{bubeck2009pure} proved that, for Bernoulli bandits, if an exploration--exploitation algorithm achieves an upper-bounded regret, it cannot reduce the expected simple regret by more than a fixed lower bound.
This establishes the fundamental difference between exploration--exploitation bandits and pure exploration bandits.
\cite{audibert2010best} identified the pure exploration problem as \textit{best arm identification} and proposed the Successive Rejects algorithm under fixed budget constraints. 
\cite{bubeck2013multiple} extended this algorithm for finding $m$-best arms and proposed the Successive Accepts and Rejects algorithm.
In another endeavour to adapt the UCB family to pure exploration scenario, the LUCB family of frequentist algorithms are proposed~\cite{kaufmann2013information}.
In the beginning, they sample all the arms. Following that, they sample both the arm with maximum expected reward and the one with maximum upper-confidence bound till the algorithm can identify each of them separately.
Existing frequentist algorithms
\cite{audibert2010best,bubeck2013multiple,kaufmann2013information} do not provide an intuitive and rigorous explanation of how a unified framework would work for both the pure exploration and the exploration--exploitation scenario.
As discussed in Section~\ref{sec:intro}, both Thompson sampling and
Bayes-UCB also lack this feature of constructing a single successful
structure for both pure exploration and exploration--exploitation.

\paragraph{Two-Phase reinforcement learning.}
Two-phase reinforcement learning problems append the exploration--exploitation problem after the pure exploration problem. 
The agent gets an initial phase of pure exploration for a given window.
In this phase, the agent collects more information about the underlying reward distributions.
Following this, the agent goes through the exploration--exploitation phase. 
In this phase, it solves the exploration--exploitation problem and focuses on maximising the cumulative reward.
This setup is perceivable as an initial online model building or `training' phase followed by an online problem solving or `testing' phase.
This problem setup often emerges in applications~\cite{faheem2015adaptive} where the decision maker explores for an initial phase to create a knowledge base and another phase to take decisions by leveraging this pre-build knowledge base. In applications, this way of beginning the exploration--exploitation is called a warm start.
Thus, two-phase reinforcement learning gives us a middle ground between model-free and model-dependent approaches in decision making which is often the path taken by a practitioner.

Formally, this knowledge-base is a prior distribution built from the agent's experience. 
Since Bayesian methods naturally accommodate and leverage prior distributions, Bayesian formulation provide the scope to approach this problem without any modification.
\cite{putta2017twophaseepisodic} approached this problem with a technique amalgamating a sampling technique, PSPE, and an extension of Thompson sampling, PSRL~\cite{osband2013psrl}, for episodic fixed horizon Markov decision processes (MDPs)~\cite{dann2015episodic}.
PSPE uses Bayesian update to create a posterior distribution for the reward distribution of a policy.
Then, PSPE samples from the distribution in order to evaluate the policies.
These two steps are performed iteratively for the initial pure exploration phase.
PSRL~\cite{osband2013psrl} is an extension of Thompson sampling for episodic MDPs.
Unlike Thompson sampling, they also use Markov chain Monte Carlo method for creating the posteriors corresponding to each of the policies.
Though the amalgamation of these two methods for the two phase problems in episodic MDPs perform reasonably, they lack a reasonable unified structure attacking the problem and a natural cause to pipeline them.

\end{toappendix}
\section{Methodology}\label{sec:method}
\textbf{Bandit Problem.}\label{sec:form}
We consider a finite number $K >1$ of independent arms.
An arm $a$ corresponds to a reward distribution $f^a_{\theta}\left( X \right)$.
We assume that the form of the probability distribution $f_{\cdot}(X)$ is known to the algorithm but the parametrisation $\theta \in \Theta$ is unknown.
We assume the reward distributions of all arms to be identical in form
but to vary over the parametrisation~$\theta$. Thus, we refer to $f^a_\theta(X)$ as $f_{\theta_a}(X)$ for specificity.
The agent sequentially chooses an arm $a_t$ at each time step $t$ that generates a sequence of rewards~$[x_t]_{t =1}^T$, where $T \in \mathbb{N}$ is the time horizon.
The algorithm computes a \textit{policy} or strategy that sequentially draws a set of arms depending on her previous actions, observations and intended goal.
The algorithm does not know the `true' parameters of the arms $\lbrace
\theta_a^{\mathrm{true}} \rbrace_{a=1}^K$ a priori.
Thus, the uncertainty over the estimated parameters $\lbrace \theta_a \rbrace_{a=1}^K$ is represented using a probability distribution $B(\theta_1, \ldots, \theta_K)$.  
We call $B(\theta_1, \ldots, \theta_K)$ the \textit{belief distribution}.
In the Bayesian approach, the algorithm starts with a prior belief distribution $B_0(\theta_1, \ldots, \theta_K)$~\cite{jaynes68priorprobabilities}.
The actions taken and rewards obtained by the algorithm till time $t$ create the history of the bandit process, $\mathcal{H}_t \triangleq \left[(a_1, x_1), \ldots, (a_{t-1}, x_{t-1}) \right]$.
This history $\mathcal{H}_t$ is used to sequentially update the belief distribution over the parameter vector as $B_t(\theta_1, \ldots, \theta_K) \triangleq \mathbb{P}(\theta_1, \ldots, \theta_K\mid \mathcal{H}_t)$.
We define the space consisting of all such distributions over $\lbrace \theta_a \rbrace_{a=1}^K$ as the \textit{belief space}~$\mathcal{B}$.
Following the stochastic bandit literature, we assume the arms to be
independent, and perform Bayesian updates of beliefs.
\setlength{\textfloatsep}{1pt}
\begin{assum}[(Independence of Arms)]\label{ass:indep}
The parameters $\lbrace \theta_a \rbrace_{a=1}^K$ are drawn independently from $K$ belief distributions $\lbrace b^a_t \left(.\right) \rbrace_{a=1}^K$, such that
$B_t(\theta_1, \ldots, \theta_K) = \prod_{a=1}^K b^a_t (\theta_a) \triangleq \prod_{a=1}^K \mathbb{P}( \theta_a\mid \mathcal{H}_t).$
\end{assum}
Though Assumption~\ref{ass:indep} is followed throughout this paper, we
  note it is not essential to develop the framework BelMan relies on,
  though it makes calculations easier.
  \begin{assum}[(Bayesian Evolution)]\label{ass:bayesian}
When conditioned over $\lbrace \theta_a \rbrace_{a=1}^K$ and the choice
    of arm, the sequence of rewards $\left[ x_1 , \ldots, x_t \right] $
    is jointly independent. Thus, the Bayesian update at the $t$-th iteration is given by 
\begin{equation}\label{eqn:beliefupdate}
b^a_{t+1} (\theta_{a}) \varpropto f_{\theta_{a}}(x_t)\times b^a_{t} (\theta_a)
\end{equation}
if $a_t=a$ and a reward $x_t$ is obtained. For all other arms, the belief remains unchanged. 
\end{assum}

\textbf{Belief-reward Manifold.} We use the joint distributions $\prob(X, \theta)$ on reward $X$ and parameter $\theta$ in order to represent the uncertainties of partial information about the reward distributions along with the stochastic nature of reward. 
  \begin{definition}[(Belief-reward distribution)]
The joint distribution $\mathbb{P}^a_t(X,\theta)$ on reward $X$ and parameter $\theta_a$ for the $a^{\mathrm{th}}$ arm at the $t^{th}$ iteration is defined as the \emph{belief-reward distribution}. 
\begin{align*}
\mathbb{P}^a_t(X,\theta) \triangleq
  \frac{b^a_t(\theta)f_{\theta}(X)}{\int\limits_{X\in \mathbb{R}}\int\limits_{\theta \in \Theta} b^a_t(\theta)f_{\theta}(X) \mathrm{d\theta}\mathrm{dx} } = \frac{1}{Z}b^a_t(\theta)f_{\theta}(X).
\end{align*}
\end{definition}
If $f_{\cdot}(X)$ is a smooth function of $\theta_a$'s, the space of all reward distributions constructs a smooth statistical manifold~\cite{amari2007infogeo}, $\mathcal{R}$.
We call $\mathcal{R}$ the \textit{reward manifold}. 
If belief $B$ is a smooth function of its parameters, the belief space $\mathcal{B}$ constructs another statistical manifold. We call $\mathcal{B}$ the \textit{belief manifold} of the multi-armed bandit process. Assumption~\ref{ass:indep} implies that the belief manifold $\mathcal{B}$ is a product of $K$~manifolds $\mathcal{B}^a \triangleq \lbrace b^a(\theta_a)\rbrace$. 
Here, $\mathcal{B}^a$ is the statistical manifold of belief distributions for the $a$th arm.
Due to the identical parametrization, the $\mathcal{B}^a$'s can be represented by a single manifold $\mathcal{B}_\theta$.
\begin{lemma}[(Belief-Reward Manifold)]\label{lemma:2}
If the belief-reward distributions $\mathbb{P}(X,\theta)$ have smooth probability density functions, their set defines a manifold $\mathcal{B}_{\theta}\mathcal{R}$ . We refer to it as the \emph{belief-reward manifold}. Belief-reward manifold is the product manifold of the belief manifold and the reward manifold, i.e. $\mathcal{B}_{\theta}\mathcal{R} = \mathcal{B}_{\theta}\times\mathcal{R}$. 
\end{lemma}
The Bayesian belief update after each of the iteration is a movement on the belief manifold from a point $b^a_t$ to another point~$b^a_{t+1}$ \emph{with maximum information gain} from the obtained reward. 
Thus, the belief-reward distributions of the played arms evolve to create a set of trajectories on the belief-reward manifold.
The goal of pure exploration is to control such trajectories collectively such that after a long enough time each of the belief-rewards accumulate enough information to resemble the `true' reward distributions well enough. 
The goal of exploration--exploitation is to gain enough information about the `true' reward distributions while increasing the cumulative reward in the path, i.e, by inducing a bias towards playing the arms with higher expected rewards.

\begin{toappendix}
  \label{app:proofs}
\subsection{KL-divergence on the Manifold.} Kullback-Liebler divergence (or KL-divergence)~\cite{kullback1997information} is a pre-metric measure of dissimilarity between two probability distributions.
  \begin{definition}[(KL-divergence)]
If there exist two probability measures $P$ and $Q$ defined over a support set $S$ and $P$ is absolutely continuous with respect to $Q$, we define the KL-divergence between them as
\begin{align*}
D_{\mathrm{KL}}(P\|Q) \triangleq \int_{S} \log\frac{\mathrm{d}P}{\mathrm{d}Q} {\mathrm{d}P}.
\end{align*}
$\frac{\mathrm{d}P}{\mathrm{d}Q}$ is the Radon-Nikodym derivative of $P$ with respect to $Q$.
\end{definition}
Since it represents the expected information lost if $P$ is encoded using $Q$, it is also called \textit{relative entropy}.
Depending on the applications, $P$ acts as the representative of `true' underlying distribution obtained from observations or data or natural law, and $Q$ represents the model or approximation of $P$.
For two probability density functions $p(s)$ and $q(s)$ defined over a support set $S$, the KL-divergence can be rewritten as
\begin{equation}\label{eqn:KL}
D_{\mathrm{KL}}(p(s)\|q(s)) = \int_{s \in S} p(s) \log\frac{p(s)}{q(s)} {ds} = - h(p(s)) + H(p(s),q(s)).
\end{equation}
Here, $h(p(s))$ is entropy of $p$ and $H(p(s),q(s))$ is the mutual information between $p$ and $q$.
Thus, from an information-theoretic perspective, we perceive
  KL-divergence as the natural divergence function on the belief-reward
  manifold when we analyse the dynamics of the entropy function on it.
Except that, any general $\alpha$-divergence function on the statistical manifold is a convex combination of $\pm 1$-divergences. Mathematically, for $\alpha \in (-1,+1)$,
\begin{align}\label{eqn:alphadiv}
\begin{split}
D^{(\alpha)}(p\|q) &\triangleq \frac{1 + \alpha}{2} D^{(+1)}(p\|q) + \frac{1 - \alpha}{2} D^{(-1)}(p\|q)\\
&= \frac{1 + \alpha}{2} D_{\mathrm{KL}}(q\|p) + \frac{1 - \alpha}{2} D_{\mathrm{KL}}(p\|q).
\end{split}
\end{align}
From a manifold perspective, it seems that the divergence function for the $\pm 1$-connections on the belief-reward manifolds and a convex mixture of $D_{\mathrm{KL}}$ divergences form the general notion of movement on any such space.
Thus, KL-divergence between two belief-reward distributions is an effective and natural quantifier of movement, and also of information accumulation during Bayesian update.
Hence, for updating the beliefs in an optimal manner, and to decrease the uncertainty, we have to represent the observations using a knowledge-base, and to minimise the KL-divergence between the knowledge-base and other distributions respectively.
If $\mathcal{P}$ are the candidate belief-reward distributions of the arms formed by accumulation of actions and rewards, and $\mathcal{Q}$ are the pseudobelief or pseudobelief-focal-reward distribution-reward distributions, the alternating minimisation scheme looks for the most succinct representation $\mathcal{Q}$ of the knowledge and the exploitation bias while choosing such arms whose belief-reward distributions resemble their true reward distributions as much as possible.

\subsection{Exponential Family}\label{sec:exp_family}
Use of KL-divergence as a divergence measure on the statistical manifolds and also the issue of representation of a random variable using sufficient statistics provoked the study of the exponential family of distributions.
Interesting properties of exponential family distributions, such as existence of finite representation of sufficient statistics, convenient mathematical form, and existence of moments, provided them a central stage in the field of mathematical statistics~\cite{brown1986expfamily}\cite{darmois1935exhaustive}\cite{kaufmann2018}\cite{koopman1936distributions}.

The \emph{exponential family}~\cite{brown1986expfamily} is a class of probability distributions which is defined by a set of \emph{natural parameters} $\omega(\theta)$ and a \emph{sufficient statistics}~$T(X)$ of the random variable $X$ as follows:
\begin{equation*}\label{eqn:exp}
	f_\theta(X) \triangleq g(X) \exp\left(\langle \omega(\theta), T(x) \rangle - A(\theta)\right).
\end{equation*}
Here, $g(X)$ is the \emph{base measure} on reward $X$ and $A(\theta)$ is called the \emph{log-partition function}.
The exponential family includes the majority of the distributions found in the bandit literature such as Bernoulli, beta, Gaussian, Poisson, exponential, and chi-squared.
For $T(X) = X$, the log-partition function is logarithm of the Laplace transform of the base measure.
\begin{example}
 Bernoulli distribution with probability of success $\theta \in (0,1)$ is defined as
 \begin{align*}
 	\centering
 	f_{\theta}(X) \triangleq \mathrm{Ber}(\theta) &= \theta^X \left(1-\theta\right)^{(1-X)}\\
 	&= \exp\left(X\log\left(\frac{\theta}{1-\theta}\right)+\log(1-\theta) \right)
 \end{align*}
 for $X \in \lbrace 0, 1\rbrace$.
 Here, the base measure $g(x)$ is $1$. The sufficient statistics is $T(X) = X$.
 The natural parameter is $\omega(\theta) = \log\left(\frac{\theta}{1-\theta}\right)$.
 The log-partition function is $A(\theta) = -\log(1-\theta) = \log(1+\exp(\omega))$.
\end{example}

We choose the exponential family to instantiate our framework not only because of its wide range and applicability but also due to its well behaving Bayesian and information geometric properties. 
%
From a sampling and uncertainty representation point of view, the exponential family is useful because of its finite representation of sufficient statistics.
Specifically, sufficient statistics of exponential family can represent any arbitrary number of independent identically distributed samples using a finite number of variables~\cite{koopman1936distributions}.
This keeps the uncertainty representation tractable for exponential family distributions.

From a Bayesian point of view, the useful property of the exponential family is the existence of \emph{conjugate distributions} which also belong to this family~\cite{brown1986expfamily}.
Two parametric distributions $f_{\bm\theta}(x)$ and
$b_{\bm\eta}(\bm\theta)$ are conjugate if the posterior distribution
$\mathbb{P}(\bm\theta|x)$ formed by multiplying them has the same form as $b_{\bm\eta}(\bm\theta)$.
Mathematically, the conjugate distribution of the distribution of Equation~\ref{eqn:exp} is given by $b_{\bm\eta}(\bm\theta) \triangleq \mathbb{P}(\bm\theta| \bm\eta, v) = f(\bm\eta, v) \exp(\langle \bm\eta, \bm\theta \rangle- vA(\bm\theta)) = f(\bm\eta, v) g(\bm\theta)^v \exp(\langle \bm\eta, \bm\theta \rangle)$.
Here, $\eta$ is the parameter of the conjugate prior and $v > 0$ corresponds to the effective number of observations that the prior contributes.
Thus, if the reward distribution belongs to the exponential family, the belief distribution is represented as:
\(
b_{\bm\eta}(\theta) \triangleq h(\theta) \exp\left(\langle \eta, T(\theta) \rangle - A(\eta)\right)
\)
with the natural parameters $\eta \in \mathbb{R}^{d'}$.
%

From information geometric point of view, exponential family distributions are flat with respect to KL-divergence~\cite{amari2007infogeo}.
Thus, both information and reverse information projections~\cite{csiszar1984iproj} that we would use in BelMan are well-defined and unique.
Thus, at each iteration, we obtain an optimal and unambiguous computation of the decision variables of BelMan.
\cite{amari2007infogeo} also stated that the necessary and sufficient condition for a parametric probability distribution to have an efficient estimator is that the distribution belongs to the exponential family and has an expectation parametrisation.
Thus, working with exponential family distributions implicitly supports the well-defined nature and possibility of getting an efficient estimation.
\end{toappendix}

\textbf{Pseudobelief: Summary of Explored Knowledge.}
In order to control the exploration, the algorithm has to construct a summary of the collective knowledge on the belief-rewards of the arms.
Since the belief-reward distribution of each arm is a point on the belief-reward manifold, geometrically their barycentre on the belief-reward manifold represents a valid summarisation of the uncertainty over all the arms~\cite{agueh2011barycenters}.
Since the belief-reward manifold is a statistical manifold, we obtain
from information geometry that this barycentre is the point on the manifold that minimises the sum of KL-divergences from the belief-rewards of all the arms~\cite{barbaresco2013information,amari2007infogeo}.
We refer to this minimising belief-reward distribution as the pseudobelief-reward distribution of all the arms.
\begin{definition}[(Pseudobelief-reward distribution)]
  A \emph{pseudobelief-reward distribution} $\bar{\mathbb{P}}_t(X,\theta)$ is a point in the belief-reward manifold that minimises the sum of KL-divergences from the belief-reward distributions $\mathbb{P}^a_t(X,\theta)$ of all the arms.
\begin{equation}\label{eqn:pb}
\bar{\mathbb{P}}_t(X,\theta) \triangleq \argmin_{\mathbb{P} \in \mathcal{B}_\theta\mathcal{R}} \sum_{a=1}^K D_{\mathrm{KL}}\left(\mathbb{P}^a_t(X,\theta)\| {\mathbb{P}}(X,\theta)\right).
\end{equation}
\end{definition}
We prove existence and uniqueness of the pseudobelief-reward for $K$
given belief-reward distributions. This proves the pseudobelief-reward to
be an unambiguous representative of collective knowledge. We also prove
that the pseudobelief-reward distribution $\bar\prob_t$ is the projection of the average belief-reward distribution $\hat{\prob}_t(X,\theta) = \sum_a \prob^a_t(X,\theta)$ on the belief-reward manifold. This result validates the claim of pseudobelief-reward as the summariser of the belief-rewards of all the arms.

\begin{toappendix}
\subsection{Pseudobelief--reward: Existence, Uniqueness and Consistency}
In order to establish pseudobelief--reward as a valid knowledge-base for all the arms, we have to prove that it exists uniquely and its parameters can be consistently estimated.

The proofs require only two assumptions.
Firstly, the belief--reward manifold can be described by a unique chart.
This implies that pdf of the belief--reward distributions is a bijective function of parameters.
Secondly, there exist unique geodesics between any two points of the belief--reward manifold.
This implies that the divergence function between any two belief--reward distributions is uniquely defined.
Instead of having such modest requirement, we represent our proofs in form of the exponential family distributions due to ease of presentation and our limited interest.
\end{toappendix}
\vspace*{-1em}
\begin{theoremrep}~\label{thm:unique}
For given set of belief-reward distributions $\lbrace \prob^a_t\rbrace_{a=1}^K$ defined on the same support set and having a finite expectation, $\bar{\prob}_t$ is
  uniquely defined, and is such that its expectation parameter verifies $\hat{\mu}_t(\theta) = \frac{1}{K} \sum_{a=1}^K \mu^a_t(\theta)$.
\end{theoremrep}\vspace*{-1em}
\begin{proof}
For belief--reward distributions $\prob^a$ and $\prob$, the KL-divergence is defined as
\begin{align*}
\kldiv{\prob^a_t}{\prob} &= \int_\theta \int_X \prob^a_t(X,\theta) \log\frac{\prob^a_t(X,\theta)}{\prob(X,\theta)} \mathrm{dx}\mathrm{d\theta}\\
&= \int_\theta \int_X f_{\theta}(X) b^a_{\xi_t}(\theta) \log\frac{b^a_{\xi_t}(\theta)}{b_{\xi}(\theta)} \mathrm{dx}\mathrm{d\theta}\\
&= \int_\theta b^a_{\xi_t}(\theta) \log\frac{b^a_{\xi_t}(\theta)}{b_{\xi}(\theta)} \left[\int_X f_{\theta}(X)\mathrm{dx}\right] \mathrm{d\theta}\\
&= \int_\theta b^a_{\xi_t}(\theta) \log\frac{b^a_{\xi_t}(\theta)}{b_{\xi}(\theta)}\mathrm{d\theta}\\
&= \mathbb{E}_{b^a_t}\left[\langle \xi^a_t, \Theta(\theta)\rangle - \Psi^a_t(\xi^a_t) - \langle \xi, \Theta(\theta)\rangle + \Psi(\xi)\right]\\
&= \langle \xi^a_t - \xi, \mu^a_t(\theta)\rangle - \Psi^a_t(\xi^a_t) +\Psi(\xi).
\end{align*}
Thus, the objective function that $\bar{\prob}$ minimises is given by
\begin{align}~\label{eqn:objfunc}
F(\prob) \triangleq \frac{1}{K} \sum_{a=1}^K \kldiv{\prob^a_t}{\prob} &= \frac{1}{K} \sum_{a=1}^K \langle \xi^a_t - \xi, \mu^a(\theta)\rangle - \frac{1}{K} \sum_{a=1}^K \Psi^a_t(\xi^a_t) + \Psi(\xi).
\end{align}
Since the exponential family distributions are dually flat~\cite{amari2007infogeo}, we get a unique expectation parametrisation $\mu(\theta)$ of the belief distributions for a given natural parametrisation $\xi$. The expectation parameter is defined as $\mu(\theta) \triangleq \mathbb{E}_{b}[\Theta(\theta)]= \nabla_\xi\Psi(\xi)$. $\mu(\theta)$ dually expresses a natural parametrisation as its dual. Mathematically, $\xi = \nabla_\mu(\langle	\xi,\mu\rangle - \Psi(\xi)) = \nabla_\mu \Phi(\mu)$. $\Psi(\xi)$ and $\Phi(\mu)$ are log-normalisers under two parametrisations and are convex conjugate to each other.
If we define $\hat{\mu}_t(\theta) \triangleq \frac{1}{K}\sum_{a=1}^K \mu^a_t$, we get a unique natural parameter $\hat{\xi}_t \triangleq \xi(\hat{\mu}_t)$. This allows us to rewrite Equation~\ref{eqn:objfunc} as
\begin{align*}
F(\prob) &=  \left[ \langle \hat{\xi}_t - \xi, \hat{\mu}_t(\theta)\rangle -\Psi(\hat{\xi}_t) + \Psi(\xi) \right]\\ 
&+ \frac{1}{K} \sum_{a=1}^K \left(\langle \xi^a_t, \mu^a_t(\theta)\rangle - \Psi^a_t(\xi)) - (\langle \xi(\hat{\mu}_t), \hat{\mu}_t(\theta)\rangle - \Psi(\xi(\hat{\mu}_t)\right)\\
&= \kldiv{\prob_{\hat{\mu}_t}}{\prob} + \frac{1}{K} \sum_{a=1}^K \Phi(\mu^a_t) - \Phi(\hat{\mu}_t) \geq \frac{1}{K} \sum_{a=1}^K \Phi(\mu^a_t) - \Phi(\hat{\mu}_t).
\end{align*}
Since $\kldiv{\prob_{\hat{\mu}_t}}{\prob} = 0$ for $\prob = \prob_{\hat{\mu}_t}$, $F(\prob)$ reaches unique minimum $F(\prob_{\hat{\mu}})$ for the belief--reward distribution with expectation parameter $\hat{\mu_t}(\theta) \triangleq \frac{1}{K}\sum_{a=1}^K \mu^a_t$.
Thus, for a given set of belief--reward distributions the pseudobelief--reward distribution $\bar{\prob}_t(X,\theta) \triangleq \prob_{\hat{\mu}_t}(X, \theta)$ is a unique distribution in belief--reward manifold.
\end{proof}

Hereby, we establish as a unique summariser of all the belief--reward distributions.
Using this uniqueness proof, we can prove that the pseudobelief--reward distribution $\bar{\prob}$ is projection of the average belief--reward distribution $\hat{\prob}$ on the belief--reward manifold.
\vspace*{-.5em}
\begin{corollaryrep}
The pseudobelief-reward distribution $\bar{\prob}_t(X,\theta)$ is the unique point on the belief-reward manifold that has minimum KL-divergence from the distribution $\hat{\prob}_t(X,\theta) \triangleq \frac{1}{K}\sum_{a=1}^K \prob^a_t(X,\theta)$.
\end{corollaryrep}\vspace*{-.5em}
\begin{proof}
KL-divergence from $\hat{\prob}_t(X,\theta)$ to any pseudobelief--reward distribution $\prob(X,\theta)$is
\begin{align*}
\kldiv{\hat{\prob}_t}{\prob} = \kldiv{\hat{\prob}_t}{\bar{\prob}_t} + \langle\hat{\xi}_t - \xi,\hat{\mu}_t\rangle - \Psi(\hat{\xi}_t) + \Psi(\xi)
= \kldiv{\hat{\prob}}{\bar{\prob}} + \kldiv{\bar{\prob}}{\prob}.
\end{align*}
Here, $\bar{\prob}_t$ is the pseudobelief distribution with $\hat{\xi}_t$ and $\hat{\mu}_t$ as defined in Theorem~\ref{thm:unique}.
Since $\hat{\prob}_t$ is a mixture of belief--reward distributions, it does not belong to the belief--reward manifold. Thus, $\hat{\prob}_t \neq \bar{\prob}_t$ and $\kldiv{\hat{\prob}_t}{\bar{\prob}_t} > 0$. Hence,  $\kldiv{\hat{\prob}_t}{\prob}$ attends unique minimum for $\prob = \bar{\prob}_t$.
\end{proof}
\begin{toappendix}
	\subsection{Focal Distribution: Visualisation}
	\begin{figure}[t!]\vspace*{-3em}
		\centering
		\includegraphics[scale=0.3]{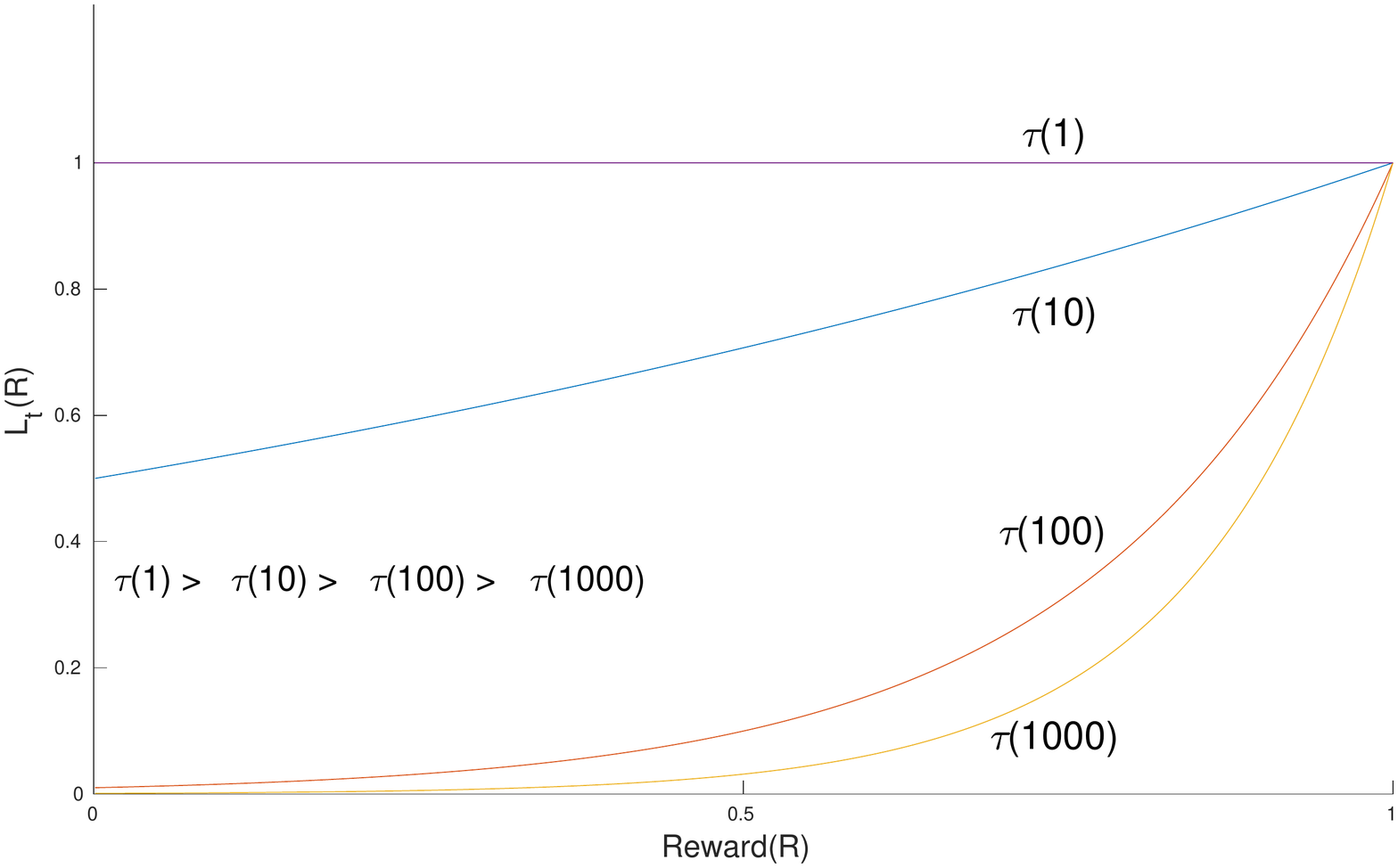}\vspace*{-3em}
		\caption{Evolution of the focal distribution over $X \in [0,1]$ for $t = 1, 10, 100$ and $1000$.}\label{fig:focal}
	\end{figure}
	The focal distribution gradually concentrates on higher rewards as the exposure $\tau(t)$ decreases with time.
	We see this feature in Figure~\ref{fig:focal}.
	Thus, it constrains using KL-divergence to choose distributions with higher rewards and induces the exploitive bias.
\end{toappendix}

\textbf{Focal Distribution: Inducing Exploitative Bias.}\label{sec:belman}
Creating a succinct pseudobelief-reward is essential for both pure exploration and exploration-- exploitation but not sufficient for maximising the cumulative reward in case of exploration--exploitation.
If a reward distribution having such increasing bias towards higher rewards is amalgamated with the pseudobelief-reward, the resulting belief-reward distribution provides a representation in the belief-reward manifold to balance the exploration--exploitation.
Such a reward distribution is called the \textit{focal distribution}.
The product of the pseudobelief-reward and the focal distribution jointly represents the summary of explored knowledge and exploitation bias using a single belief-reward distribution. We refer to this as the \textit{pseudobelief-focal-reward distribution-reward distribution}
In this paper, we use $\exp\left(\frac{X}{\tau(t)}\right)$ with a time dependent and controllable parameter $\tau(t)$ as the reward distribution inducing increasing exploitation bias.
\begin{definition}[(Focal Distribution)]
A \emph{focal distribution} is a reward distribution of the form $L_t(X) \propto \exp\left(\frac{X}{\tau(t)}\right)$, where $\tau(t)$ is a decreasing function of $t \geq 1$. We term $\tau(t)$ the \emph{exposure} of the focal distribution.
\end{definition} \vspace{-.5em}
Thus, the pseudobelief-focal-reward distribution-reward distribution is represented as \(\bar{\mathbb{Q}}(X,\theta)\triangleq \frac{1}{\bar{Z}_t}\bar{\mathbb{P}}(X,\theta)\exp\left(\frac{X}{\tau(t)}\right)\), where the normalisation factor $\bar{Z}_t = \int_{X \in \mathbb{R}}\int_{\theta \in \Theta}
\bar{\mathbb{P}}(X,\theta)\exp\left(\frac{X}{\tau(t)}\right)\mathrm{d\theta}\mathrm{dx}$. Following Equation~\eqref{eqn:pb}, we compute the pseudobelief-focal-reward distribution as  \vspace{-.7em}
\begin{equation*} \vspace{-.7em}
	\bar{\mathbb{Q}}_t(X,\theta) \triangleq \argmin_{\bar{\mathbb{Q}}} \sum_{a=1}^K D_{\mathrm{KL}}\left(\mathbb{P}^a_{t-1}(X, \theta)\| \bar{\mathbb{Q}}(X,\theta)\right).
\end{equation*}
The focal distribution gradually concentrates on higher rewards as the exposure $\tau(t)$ decreases with time.
Thus, it constrains using KL-divergence to choose distributions with higher rewards and induces the exploitive bias.
From Theorem~\ref{thm:consist}, we obtain $\frac{1}{\tau(t)}$ has to grow in the order $\Omega(\frac{1}{\sqrt{t}})$ for exploration--exploitation bandit problem independent of the family of reward distribution.
Following the bounds obtained in~\cite{garivier2011klucb}, we set the exposure  $\tau(t) = [\log(t)+C\times\log(\log(t))]^{-1}$ for experimental evaluation, where $C$ is a constant (we choose the value $C=15$ in the experiments) . As the exposure $\tau(t)$ decreases with $t$, the focal distribution gets more concentrated on higher reward values. For the pure exploration bandits, we set the exposure $\tau(t)=\infty$ to remove any bias towards higher reward values i.e, exploitation.

\setlength{\textfloatsep}{1pt}
\begin{algorithm}[t!]
\begin{algorithmic}[1]
   \STATE {\bfseries Input:} Time horizon $T$, Number of arms $K$, Prior on parameters $B_0$, Reward function $f$, Exposure $\tau(t)$.
   \FOR{$t=1$ {\bfseries to} $T$}
   \STATE{$/*$ \textit{I-projection} $*/$}
   \STATE Draw arm $a_t$ such that
   \vspace{-.75em}
   \[\hspace*{-0.2cm}
    a_t = \argmin_{a} D_{\mathrm{KL}}\left(\mathbb{P}^a_{t-1}(X,\theta)\|\bar{\mathbb{Q}}_{t-1}(X,\theta)\right).
     \]
   \vspace{-1em}
   \STATE{$/*$ \textit{Accumulation of observables} $*/$}
   \STATE Sample a reward $x_t$  out of $f_{\theta_{a_t}}$.
   \STATE Update the belief-reward distribution of $a_t$ to $\mathbb{P}^a_{t}(X, \theta)$ using Bayes' theorem.   
   \STATE{$/*$ \emph{Reverse I-projection} $*/$} 
   \STATE Update the pseudobelief-reward distribution to
   \vspace{-.75em}
   \[\hspace*{-0.7cm}
   \bar{\mathbb{Q}}_{t}(X,\theta) =\argmin_{\bar{\mathbb{Q}} \in \mathcal{B}_\theta\mathcal{R}} \sum_{a = 1}^K D_{\mathrm{KL}}\left(\mathbb{P}^a_t(X,\theta)\|\bar{\mathbb{Q}}(X,\theta)\right).
   \]
   \vspace{-1em}
   \ENDFOR
\end{algorithmic}
\caption{BelMan}\label{alg:belman}
\end{algorithm}

\textbf{BelMan: An Alternating Projection Scheme.}\label{sec:scheme}
A bandit algorithm performs three operations in each step-- chooses an arm, samples from the reward distribution of the chosen arm and incorporate the sampled reward to update the knowledge-base.
BelMan (Algorithm~\ref{alg:belman}) performs the first and the last operations by alternately minimising the KL-divergence $D_{\mathrm{KL}}(.\|.)$~\cite{kullback1997information} between the belief-reward distributions of the arms and the pseudobelief-focal-reward distribution-reward distribution. BelMan chooses to play the arm whose belief-reward incurs minimum KL-divergence with respect to the pseudobelief-focal-reward distribution. Following that, BelMan uses the reward collected from the played arm to do Bayesian update of the belief-reward and to update the pseudobelief-focal-reward distribution-reward distribution to the point minimising the sum of KL-divergences from the belief-rewards of all the arms.
\cite{csiszar1984iproj} geometrically formulated such minimisation of KL-divergence with respect to a participating distribution as a projection to the set of the other distributions.
For a given $t$, the belief-reward distributions of all the arms $\mathbb{P}_t^a(X,\bm\theta)$ form a set $\mathcal{P}\subset \mathcal{B_{\theta}R}$ and the pseudobelief-focal-reward distribution-reward distributions $\bar{\mathbb{Q}}_t(X,\bm\theta)$ constitute another set $\mathcal{Q} \subset \mathcal{B_{\theta}R}$.


\begin{toappendix}
\subsection{Condition for Existence of Alternating Projection Scheme}
Both I- and rI-projections are valid and well-defined if the
  KL-divergence between any two distributions in $\mathcal{P}$ and
  $\mathcal{Q}$ is defined and finite.
  \begin{assum}[(Absence of singularities)]\label{ass:supp2}
The distribution families $\mathcal{P}$ and $\mathcal{Q}$ are defined over the sets $Supp(\mathcal{P}) \triangleq \lbrace a : p(a) > 0, \forall p \in \mathcal{P} \rbrace$ and $Supp(\mathcal{Q}) \triangleq \lbrace a : q(a) > 0, \forall p \in \mathcal{P} \rbrace$ respectively.
Moreover, none of the supports are empty and $Supp(\mathcal{P}) \subseteq Supp(\mathcal{Q})$.
\end{assum}
\subsection{Implications of Alternating Projections}
\end{toappendix}
\vspace*{-1em}
\begin{definitionrep}[(I-projection)]\label{def:iproj}
  The \emph{information projection} (or \emph{I-projection}) of a
  distribution $\bar{\mathbb{Q}} \in \mathcal{Q}$ onto a non-empty, closed, convex set
  $\mathcal{P}$ of probability distributions, $\prob^a$'s, defined on a fixed support set is
  defined by the probability distribution ${\prob^a}^* \in \mathcal{P}$ that has
  minimum KL-divergence to $q$:
  \({\prob^a}^* \triangleq \argmin_{\prob^a \in \mathcal{P}} D_{\mathrm{KL}} (\prob^a\| \bar{\mathbb{Q}}).
  \)
\end{definitionrep}
BelMan decides which arm to pull by an
I-projection of the pseudobelief-focal-reward distribution onto the beliefs-rewards of each of the arms (Lines 3--4).
This operation amounts to computing \vspace*{-.6em}
\begin{align*}
a_t &\triangleq \argmin_a D_{\mathrm{KL}}\left(\mathbb{P}^a_{t-1}(X, \theta)\|\bar{\mathbb{Q}}_{t-1}(X,\theta)\right)\\ \hspace*{-1.5em}&= \argmax_a \left( \mathbb{E}_{\mathbb{P}^a_{t-1}(X,\theta)}\left[\frac{X}{\tau(t)} \right] - D_{\mathrm{KL}}\left(b^a_{t-1}(\theta)\|b_{\bar{\bm\eta}_{t-1}}(\theta)\right)\right)\vspace*{-2em}
\end{align*}
The first term symbolises the expected reward of arm $a$.
Maximising this term alone is analogous to greedily exploiting the present information about the arms.
The second term quantifies the amount of uncertainty that can be decreased if arm $a$ is chosen on the basis of the present pseudobelief.
The exposure $\tau(t)$ of the focal distribution keeps a weighted balance between exploration and exploitation.
Decreasing $\tau(t)$ decreases the exploration with time which is quite an intended property of an exploration--exploitation algorithm.

\begin{toappendix}
Since $D_{\mathrm{KL}}(p(s)\|q(s)) = - h(p(s)) + H(p(s),q(s))$, we
observe that the I-projection $p^*$ is the distribution in $\mathcal{P}$
that maximises the entropy $h(p)$ of $\mathcal{P}$, while
minimising the mutual information $H(p,q)$: it is the distribution in $\mathcal{P}$ which is most similar to $q$.
This implies that the I-projection $p^*$ captures at least the first
moment, i.e., the expectation of the fixed distribution $q$.

  In the last part (Lines 8--9), the updated beliefs are used to obtain
the pseudobelief-focal-reward distribution using rI-projection. Following
  Theorem~\ref{thm:unique}, rI-projection would lead to a unique
  pseudobelief-focal-reward distribution for a given set of belief-rewards and
  exposure $\tau(t)$.
Here, BelMan is inducing the exploitative bias. It keeps the
  pseudobelief-focal-reward distribution away from the `actual' barycentre of
  the belief-reward distributions and pushes it towards the arms with
  higher expected reward. Increasing exploitative bias eventually merges
the pseudobelief-focal-reward distribution to the distribution of the arm having
the highest expected reward. 
\end{toappendix}

Following that (Line 5--7), the agent plays the chosen arm $a_t$ and samples a reward $x_t$.
This observation is incorporated in the belief of the arm using Bayes' rule of Equation~\eqref{eqn:beliefupdate}.\vspace*{-.6em}
\begin{definitionrep}[(rI-projection)]\label{def:reviproj}
  The \emph{reverse information projection} (or \emph{rI-projection}) of a
  distribution $\prob^a \in \mathcal{P}$ onto $\mathcal{Q}$, which is also a
  non-empty, closed, convex set of probability distributions on a fixed
  support set, is defined by the distribution $\bar{\mathbb{Q}}^* \in \mathcal{Q}$ that
  has minimum KL-divergence from $\prob^a$:
$\bar{\mathbb{Q}}^* \triangleq \argmin_{\bar{\mathbb{Q}} \in \mathcal{Q}}$ $D_{\mathrm{KL}} (\prob^a \|\bar{\mathbb{Q}}).$
\end{definitionrep}
\vspace*{-.5em}
\begin{toappendix}
The rI-projection finds the distribution $q^{\ast}$ from a space of candidate distributions $\mathcal{Q}$ that encodes maximum information of the distribution $p$.
If the set of candidate distributions is engendered by a statistical
model,  the rI-projection of the empirical distribution formed from
samples to the model is equivalent to finding the \emph{maximum likelihood estimate}.
Since rI-projection aims to maximise the complete likelihood rather than finding a distribution with similar entropy, $q^*$ also captures higher moments of the fixed distribution $p$.
Thus, it is computationally more demanding but more informative than I-projection.

Due to the underlying minimisation operation, if we begin from $p_0 \in
\mathcal{P}$ and $q_0 \in \mathcal{Q}$ and alternately perform
I-projection and reverse I-projection, it will lead to two distributions
$p^*$ and $q^*$ for which the KL-divergence between sets~$\mathcal{P}$
and $\mathcal{Q}$ are minimum~\cite{csiszar1984iproj}.
\end{toappendix}
\begin{toappendix}
\subsection{Law of Convergence for the Pseudobelief-reward Distribution}
We are simultaneously approximating the belief--reward parameters as well as the pseudobelief--reward parameters.
If we look into the belief update step (Equation~\ref{eqn:beliefupdate}), we observe that the belief distribution of each arm $b^a_{\xi_t}(\theta)$ is updated by incorporating i.i.d samples obtained from the reward distribution of that arm.
Let us assume that BelMan has played total $T$ times and any arm $a$ for $t^a_T$ times.
Since we are doing na\"{i}ve Bayesian updates with i.i.d. samples, the belief distributions will follow central limit theorem.
This means that if $\tilde{\mu}^a_{t^a}$ is the estimate of the expectation parameters of the belief distribution of arm $a$ constructed from samples $\lbrace X^a_i\rbrace_{i=1}^{t^a_T}$, $\sqrt{t^a_T}(\tilde{\mu}^a_{t^a_T} - \mu^a)$ converges in distribution to a centered normal random vector in $\mathcal{N}(0,\Sigma^a)$.
In Theorem~\ref{thm:clt}, we show that the estimator of the mean parameters of pseudobelief is also consistent with these estimators and satisfies central limit theorem.
\end{toappendix}
\vspace*{-1em}
\begin{theoremrep}[(Central limit theorem)]~\label{thm:clt}
If $\tilde{\bar{\mu}}_{T} \triangleq \frac{1}{K}\sum_{a=1}^K \tilde{\mu}^a_{t^a_T}$ is estimator of the expectation parameters of the pseudobelief distribution, $\sqrt{T}(\tilde{\bar{\mu}}_{T} - \bar{\mu})$ converges in distribution to a centered normal random vector in $\mathcal{N}(0,\bar{\Sigma})$. The covariance matrix $\bar{\Sigma} = \sum_{a=1}^K \lambda_a \Sigma^a$ such that $\frac{T}{K^2 t^a_T}$ tends to $\lambda^a$ as $T \rightarrow \infty$.
\end{theoremrep}\vspace*{-1em}
\begin{proof}
The characteristics function for $\sqrt{N}(\tilde{\bar{\mu}}_{N} - \bar{\mu})$ is
\begin{align*}
\bm\Phi_{\sqrt{T}(\tilde{\bar{\mu}}_{T} - \bar{\mu})}(t) &= \mathbb{E}\left[\exp(\iota\langle t, \sqrt{T}(\tilde{\bar{\mu}}_{T} - \bar{\mu})\rangle)\right]\\
&= \mathbb{E}\left[\exp(\iota\langle t, \frac{\sqrt{T}}{K}\sum_{a=1}^K (\tilde{\mu}^a_{t^a_T} - \mu^a)\rangle)\right]\\
&= \prod_{a=1}^K\mathbb{E}\left[\exp(\iota\langle t, \frac{\sqrt{T}}{K} (\tilde{\mu}^a_{t^a_T} - \mu^a)\rangle)\right]\\
&= \prod_{a=1}^K\mathbb{E}\left[\exp(\iota\langle \frac{\sqrt{T}}{K\sqrt{t^a_T}}t,  \sqrt{t^a_T}(\tilde{\mu}^a_{t^a_T} - \mu^a)\rangle)\right]\\
&= \prod_{a=1}^K \bm\Phi_{\sqrt{t^a_T}(\tilde{\mu}^a_{t^a_T} - \mu^a)}\left(\frac{\sqrt{T}}{K\sqrt{t^a_T}}t\right)
\end{align*}
Since each of the $\sqrt{t^a_T}(\tilde{\mu}^a_{t^a_T} - \mu^a)$ converges in distribution to a random vector that follows $\mathcal{N}(0,\Sigma^a)$, the covariance matrix for $\sqrt{T}(\tilde{\bar{\mu}}_{T} - \bar{\mu})$ would be $\lim_{T\rightarrow\infty} \sum_{a=1}^K \left(\frac{\sqrt{T}}{K\sqrt{t^a_T}}\right)^2 \Sigma^a = \sum_{a=1}^K \lambda^a \Sigma^a \triangleq \bar{\Sigma}$.
\end{proof}

Theorem~\ref{thm:clt} shows that the parameters of pseudobelief can be constantly estimated and their estimation would depend on the accuracy of the estimators of individual arms with a weight on the number of draws on the corresponding arms.
Thus, the uncertainty in the estimation of the parameter is more influenced by the arm that is least drawn and less influenced by the arm most drawn.
In order to decrease the uncertainty corresponding to pseudobelief, we have to draw the arms less explored.

We need an additional assumption before moving into the asymptotic consistency claim in Theorem~\ref{thm:consist}.
\begin{assum}[Bounded log-likelihood ratios]\label{prop:concofbelief}
The log-likelihood of the posterior belief distribution at time $t$ with respect to the true posterior belief distribution is bounded such that $\lim_{t \rightarrow \infty} \abs{\log \frac{\mathbb{P}^a (X, \theta)}{\mathbb{P}^a_t (X, \theta)}} \leq C < \infty$ for all $a$.
\end{assum}
This assumption helps to control the convergence of sample KL divergences in to the true KL-divergences as the number of samples grow infinitely. This is a relaxed version of Assumption 2 employed in~\cite{gopalan2015thompson} to bound the regret of Thompson sampling. This is also often used in the statistics literature to control the convergence rate of posterior distributions~\cite{shen2001rates}\cite{wong1995}.

\begin{toappendix}
\subsection{Proof of Theorem~\ref{thm:consist}}
\end{toappendix}
\vspace*{-1em}
\begin{theoremrep}[(Asymptotic consistency)]\label{thm:consist}
Given ${\tau(t)} = \frac{1}{\log{t}+ c\times \log{\log{t}}}$ for any $c \geq 0$, BelMan will asymptotically converge to choosing the optimal arm in case of a bandit with bounded reward and finite arms. Mathematically, if there exists $\mu^*  \triangleq \max_{a} \mu(\theta_a)$,\vspace*{-.7em}
\begin{equation}
\lim_{T \rightarrow \infty} \frac{1}{T} ~\mathbb{E}\left[\sum_{t=1}^T ~X_{a_t}\right] = \mu^*.
\end{equation}\vspace*{-1.5em}
\end{theoremrep}
\begin{toappendix}
We reformulate this result more precisely using Lemma~\ref{lemma:3}.
\begin{lemma}~\label{lemma:3}
 If Assumption~\ref{prop:concofbelief} is true and there exists at least an optimal arm with expected reward $\mu^* \triangleq \max_{a} \mu(\theta_a)$, and the exposure satisfies
  $\lim_{t\rightarrow \infty}\tau(t) \leq \frac{1}{\sqrt{2}C}$, then BelMan would satisfy asymptotic consistency
\begin{equation}
\lim_{T \rightarrow \infty} \frac{1}{T} \mathbb{E}\left[\sum_{t=1}^T \left(X_{A_t}\right)\right] = \mu^*.
\end{equation}
\end{lemma}

\begin{proof}
Without loss of generality, let us consider that there exists at least one optimal arm and it is identified as the arm $a=1$. 
At the I-projection step, we choose the arm that has minimum KL-divergence $\kldiv{\prob^a_t(X,\theta)}{\bar{\mathbb{Q}}(X, \theta)}$ from the pseudobelief--focal distribution.
Thus, we have to prove that for large $t$ and for all $a \neq 1$, \[ \lim_{t \rightarrow \infty} \prob(\kldiv{\prob^1_t(X,\theta)}{\bar{\mathbb{Q}}(X, \theta)} - \kldiv{\prob^a_t(X,\theta)}{\bar{\mathbb{Q}}(X, \theta)} < 0) = 1.\]
This is equivalent to proving that almost surely
\begin{equation}\label{eqn:nonneg}
    \lim_{t \rightarrow \infty} \left(\kldiv{\prob^1_t(X,\theta)}{\bar{\mathbb{Q}}(X, \theta)} - \kldiv{\prob^a_t(X,\theta)}{\bar{\mathbb{Q}}(X, \theta)}\right) < 0.
\end{equation}
We begin as follows,
\begin{align*}
&\kldiv{\prob^1_t(X,\theta)}{\bar{\mathbb{Q}}(X, \theta)} - \kldiv{\prob^a_t(X,\theta)}{\bar{\mathbb{Q}}(X, \theta)}\\
= &\underbrace{\int_{X}\int_{\theta} \mathbb{P}^1_t(X,\theta)
  \log{\prob^1_t}(X, \theta)~\mathrm{d\theta}~\mathrm{dX} -
  \int_{X}\int_{\theta} \mathbb{P}^a_t(X,\theta) \log{\prob^a_t}(X,
  \theta)~\mathrm{d\theta}~\mathrm{dX}}_{\mathbf{T1}} \\& + \underbrace{\int_{X}\int_{\theta} \left[\mathbb{P}^a_t(X,\theta) - \mathbb{P}^1_t(X,\theta) \right] \log\bar{\mathbb{Q}}(X, \theta)~\mathrm{d\theta}~\mathrm{dX}}_{\mathbf{T2}}
\end{align*}
The first term $\mathbf{T1}$ is the difference in entropy in two of the arms. 
\begin{align*}
\mathbf{T1} = &\int_{X}\int_{\theta} \mathbb{P}^1_t(X,\theta) \log{\prob^1_t}(X, \theta)~\mathrm{d\theta}~\mathrm{dX} - \int_{X}\int_{\theta} \prob^a_t(X,\theta) \log{\prob^a_t}(X, \theta)~\mathrm{d\theta}~\mathrm{dX}\\
= &\int_{X}\int_{\theta} \left[\prob^a_t(X,\theta) - \prob^1_t(X,\theta) \right] \log\prob^1_t(X, \theta)~\mathrm{d\theta}~\mathrm{dX}~-~ \kldiv{\prob^a_t(X,\theta)}{\prob^1_t(X, \theta)}\\
\underset{(a)}{\leq} &\int_{X}\int_{\theta} \left[\prob^a_t(X,\theta) - \prob^1_t(X,\theta) \right] \log\prob^1_t(X, \theta)~\mathrm{d\theta}~\mathrm{dX}\\
\underset{(b)}{\leq} &\int_{X}\int_{\theta} \abs{\left[\mathbb{P}^a_t(X,\theta) - \mathbb{P}^1_t(X,\theta) \right] \log\mathbb{P}^1_t(X, \theta)}~\mathrm{d\theta}~\mathrm{dX}\\
\underset{(c)}{\leq} &\sup_{X, \theta} \abs{\log\mathbb{P}^1_t(X, \theta)} \int_{X}\int_{\theta} \abs{\mathbb{P}^a_t(X,\theta) - \mathbb{P}^1_t(X,\theta)}~\mathrm{d\theta}~\mathrm{dX}\\
\underset{(d)}{\leq} & \sup_{X, \theta} \abs{\log\mathbb{P}^1_t(X, \theta)}\sqrt{\frac{\log 2}{2}~\kldiv{\prob^a_t(X,\theta)}{\prob^1_t(X, \theta)}}\\
\end{align*}
The inequality $(a)$ is due to the non-negativity of KL-divergence.
Inequality $(b)$ is derived from the monotonicity of integrals. This means that if $f \leq g$ for all $w \in W$ then $\int_{w \in W} f(w)~\mathrm{dw} ~\leq \int_{w \in W}~g(w)~\mathrm{dw}$.
Boundedness of the logarithmic density function of the pseudobelief-reward as stated in Proposition~\ref{prop:concofbelief} results to inequality $(c)$.
Inequality $(d)$ is obtained from Pinsker's inequality~\cite{cover2012elements}. 

Similarly, we get for the second term $\mathbf{T2}$:
\begin{align*}
\mathbf{T2} &= \int_{X}\int_{\theta} \left[\mathbb{P}^a_t(X,\theta) - \mathbb{P}^1_t(X,\theta) \right] \log\bar{\mathbb{Q}}(X, \theta)~\mathrm{d\theta}~\mathrm{dX})\\
&= \int_{X}\int_{\theta} \left[\mathbb{P}^a_t(X,\theta) - \mathbb{P}^1_t(X,\theta) \right] \log\left( \prod_a {\prob^a_t(X, \theta)}^{\lambda^a_t}\right)~\mathrm{d\theta}~\mathrm{dX}\\ &- \frac{1}{\tau(t)}\mathbb{E}_{\mathbb{P}^1_t(X,\theta)- \mathbb{P}^a_t(X,\theta)} \left[X \right] +\log\bar{Z}_t \times \mathbb{E}_{\mathbb{P}^1_t(X,\theta)- \mathbb{P}^a_t(X,\theta)} \left[1\right]\\
&\underset{(e)}{\leq} \sup_{X, \theta} \abs{\log\mathbb{P}^1_t(X, \theta)} \sqrt{\frac{\log 2}{2}}~\sqrt{\kldiv{\mathbb{P}^a_t(X,\theta)}{\mathbb{P}^1_t(X, \theta)}} - \frac{\Delta^a_t}{\tau(t)}.
\end{align*}
Here, $\Delta^a_t \triangleq \mu^1_t -\mu^a_t$, which means the difference between the expected reward of the optimal arm and the suboptimal arm $a$.
Inequality $(e)$ is obtained by applying AM-GM inequality, inequalities $(a)$, $(b)$, $(c)$, and $(d)$ in sequence.
Thus, 
\begin{align*}
    \mathbf{T1}+\mathbf{T2} &\leq \sup_{X, \theta} \abs{\log\mathbb{P}^1_t(X, \theta)}\sqrt{2\log 2} \sqrt{\kldiv{\mathbb{P}^a_t(X,\theta)}{\mathbb{P}^1_t(X, \theta)}} - \frac{\Delta^a_t}{\tau(t)}\\
    &= \sqrt{2\log 2} \sqrt{\kldiv{\mathbb{P}^a_t(X,\theta)}{\mathbb{P}^1_t(X, \theta)}}\\ &\left(  \sup_{X, \theta} \abs{\log\mathbb{P}^1_t(X, \theta)} - \frac{1}{\tau(t)} \frac{\Delta^a_t}{\sqrt{\kldiv{\mathbb{P}^a_t(X,\theta)}{\mathbb{P}^1_t(X, \theta)}}} \right)\\
    &\leq \sqrt{2\log 2} \sqrt{\kldiv{\mathbb{P}^a_t(X,\theta)}{\mathbb{P}^1_t(X, \theta)}} \left(  \sup_{X, \theta} \abs{\log\mathbb{P}^1_t(X, \theta)} - \frac{1}{\sqrt{2}\tau(t)}\right)
\end{align*}
If we consider $\lim_{t \rightarrow \infty}$ for both sides of the inequality, we observe Equation~\ref{eqn:nonneg} is true if $$\lim_{t\rightarrow \infty} \left(\sup_{X, \theta} \abs{\log\mathbb{P}^1_t(X, \theta)} - \frac{1}{\sqrt{2}\tau(t)} \right) < 0.$$ This holds as $\kldiv{\mathbb{P}^a_t(X,\theta)}{\mathbb{P}^1_t(X, \theta)} > 0$ for all $a$ and $t$.
By Assumption 4, we get $\lim_{t\rightarrow \infty} \sup_{X, \theta} \abs{\log\mathbb{P}^1_t(X, \theta)} \leq C + \log\mathbb{P}^1_t(X, \theta) = C'$(say).
Thus, we get in order to satisfy the inequality $\lim_{t \rightarrow \infty} \tau(t) < \frac{1}{\sqrt{2}C'}$ which is in our premise.
\end{proof}
\begin{lemma}~\label{lemma:4}
 For $\tau(t) = \frac{1}{\log{t}+c\times\log{\log{t}}}$ with $c \geq 0$, $\lim_{t\rightarrow \infty} \tau(t) < \frac{1}{C}$ for any $C < \infty$.
\end{lemma}
\begin{proof}
Since $\lim_{t \rightarrow \infty} \frac{1}{\log{t}+c\log{\log{t}}} = 0$, the aforementioned claim holds true.
\end{proof}
Lemma~\ref{lemma:3} and~\ref{lemma:4} together prove Theorem~\ref{thm:consist}. This proves that BelMan is asymptotically consistent for finite-arm stochastic bandit problems.

\end{toappendix}

We intuitively validate this claim.
We can show the KL-divergence between belief-reward of arm $a$ and the pseudobelief-focal-reward is $D_\mathrm{KL}(\mathbb{P}^a_t(X,\theta)\|\bar{\mathbb{Q}}(X, \theta))$ $= (1- \lambda^a) h(b^a_t) - \frac{1}{\tau(t)} \mu^a_t$, for $\lambda^a$ computed as per Theorem~\ref{thm:clt}. Here, $h(b^a_t)$ denotes the entropy of belief distribution $b^a_t$ of arm $a$ at time $t$.
As $t \rightarrow \infty$, the entropy of belief on each arm reduces to a constant dependent on its internal entropy.
Thus, when $\frac{1}{\tau(t)}$ exceeds the entropy term for a large $t$, BelMan greedily chooses the arm with highest expected reward.
Hence, BelMan is asymptotically consistent.

\begin{figure*}[t!]\vspace*{-.8em}
\hspace{-1cm} \includegraphics[scale=0.35,trim={6cm 2cm 3.5cm 10.5cm},clip]{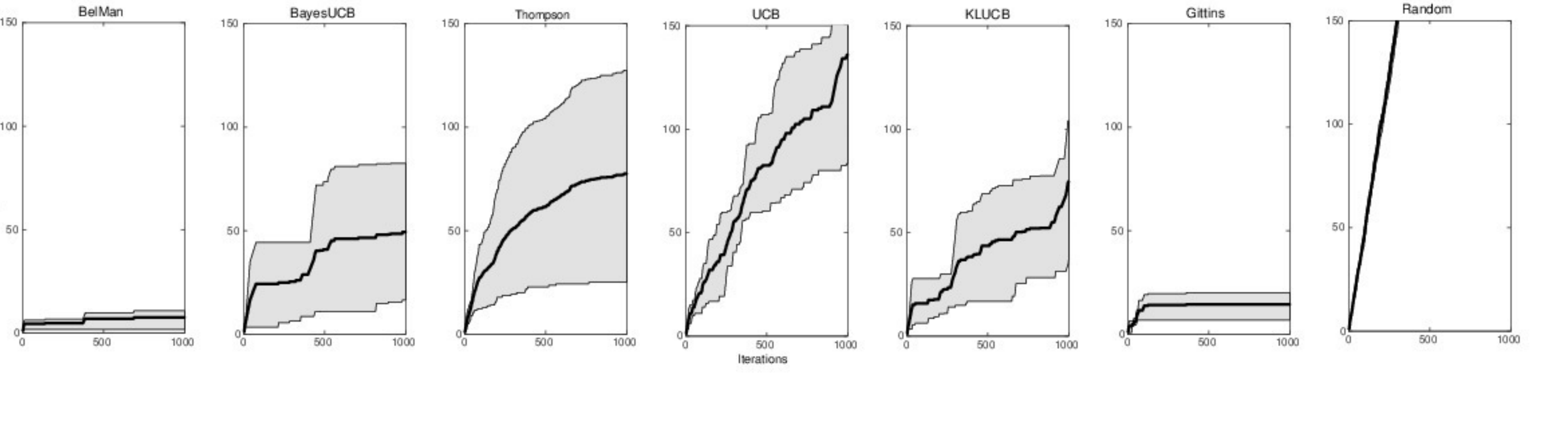}\vspace*{-1cm}
\caption{Evolution of number of suboptimal draws for 2-arm Bernoulli bandit with expected rewards 0.8 and 0.9 for 1000 iterations. The dark black line shows the average over 25 runs. The grey area shows the 75 percentile.}\label{fig:ber_2_80_90}
\hspace{-1cm}\includegraphics[scale=0.35,trim={6cm 2cm 3.5cm 11cm},clip]{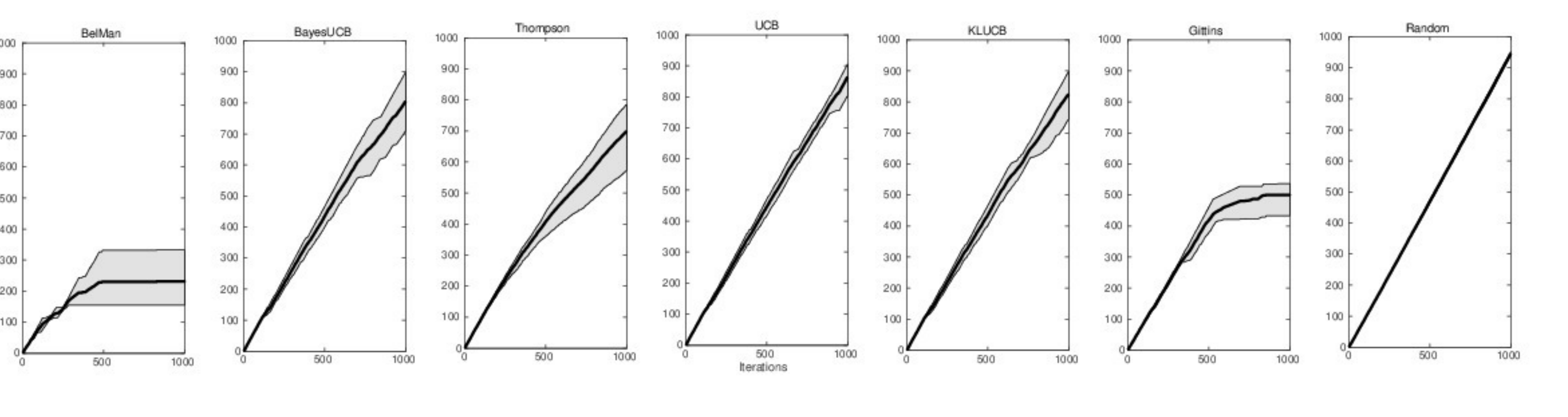}\vspace*{-1cm}
\caption{Evolution of number of suboptimal draws for 20-arm Bernoulli bandit with expected rewards [0.25 0.22 0.2 0.17 0.17 0.2 0.13 0.13 0.1 0.07 0.07 0.05 0.05 0.05 0.02 0.02 0.02 0.01 0.01 0.01] for 1000 iterations.}\label{fig:ber_20}
\hspace*{-1.4cm}
\includegraphics[scale=0.35,trim={5cm 2cm 4cm 12.5cm},clip]{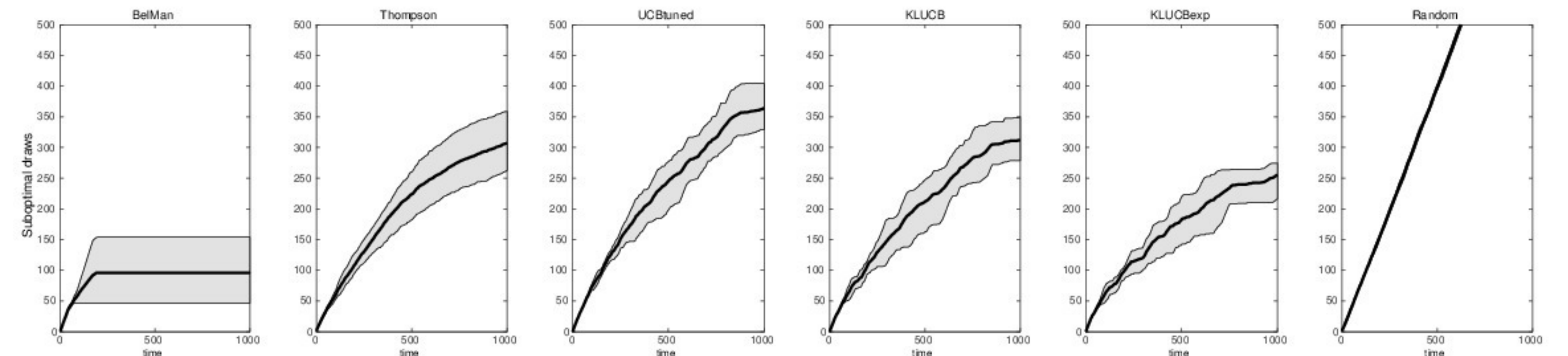}\vspace*{-0.6cm}
\caption{Evolution of number of suboptimal draws for 5-arm bounded exponential bandit with expected rewards 0.2, 0.25, 0.33, 0.5, and 1.0 for 1000 iterations.}\label{fig:exp_5}
\end{figure*}

BelMan is applicable to any belief-reward distribution for which KL-divergence is computable and finite. Additionally for reward distributions belonging to the exponential family of distributions, the belief distributions, being conjugate to the reward distributions, also belong to the exponential family~\cite{brown1986expfamily}. This makes belief-reward distributions flat with respect to KL-divergence. Thus, both
I-and rI-projections in BelMan are well-defined and unique for exponential family reward distributions. Furthermore, if we identify the belief-reward distributions with expectation parameters, we obtain the pseudobelief as an affine sum of them. This allows us to compute belief-reward distribution directly instead of computing its dependence on each belief-reward separately. The exponential family includes the majority of the distributions found in the bandit literature such as Bernoulli, beta, Gaussian, Poisson, exponential, and~$\chi^2$.
\begin{toappendix}
	For exploration--exploitation bandit problem, we observe that $\tau(t)$ has to be a positive valued function of time $t$ that asymptotically decreases with time.
	Such decay in the value of exposure $\tau(t)$ adaptively increases the importance of reward maximisation over minimising the KL-divergence between the belief-reward of selected arm and the pseudobelief-reward.
	This mechanism allows BelMan to adaptively balance between the exploration and exploitation components.
	
	The growth rate proposed for exposure, $O(\frac{1}{\log{t}})$, is a loose bound. Beside this, it is also distribution independent. Thus, we observe a gap between the bound on exposure growth obtained here, and the one used in practice. It would be interesting to find out tighter bounds with more specific constants for given reward distributions. 
\end{toappendix}

\begin{toappendix}
\subsection{BelMan for Exponential Family Distributions}\label{sec:expfamily}
As mentioned in Section B.2, \emph{exponential family}~\cite{brown1986expfamily} is a class of
probability distributions which can be defined using a set of
\emph{natural parameters} $\omega(\theta)$ and a given natural
\emph{sufficient statistics}~$T(X)$ as follows:
\begin{equation*}
f_\theta(X) \triangleq h(X) \exp\left(\langle \omega(\theta), T(X) \rangle - A(\theta)\right).
\end{equation*}
Here, $h(X)$ is the \emph{base measure} on reward $X$ and $A(\theta)$ is called the \emph{log-partition function}.
The exponential family includes the majority of the distributions found in the bandit literature such as Bernoulli, beta, Gaussian, Poisson, exponential, and chi-squared.

We choose the exponential family to instantiate our framework not only because of its wide range and applicability but also due to its well behaving Bayesian and information geometric properties. 
%
From a Bayesian point of view, the most useful property of the exponential family is the  existence of \emph{conjugate distributions} which also belong to this family~\cite{brown1986expfamily}.
Two parametric distributions $f_{\bm\theta}(X)$ and
$b_{\bm\eta}(\bm\theta)$ are conjugate if the posterior distribution
$\mathbb{P}(\bm\theta|X)$ formed by multiplying them has the same form as $b_{\bm\eta}(\bm\theta)$.
Thus, if the reward distribution belongs to the exponential family, the
belief distribution is represented as:
\(
b_{\bm\eta}(\theta) \triangleq h(\theta) \exp\left(\langle \bm\eta, T(\theta) \rangle - A(\bm\eta)\right)
\)
with the natural parameters $\bm\eta$.
%

Since exponential family distributions are flat with respect to KL-divergence \cite{amari2007infogeo}, both I-and rI-projections in BelMan are well-defined and unique.
Thus, at each iteration, we obtain an optimal and unambiguous choice of the arm and pseudobelief respectively.
\cite{amari2007infogeo} also stated that the necessary and sufficient condition for a parametric probability distribution to have an efficient estimator is that the distribution belongs to the exponential family and has an expectation parametrisation.
Thus, working with exponential family distributions implicitly supports the well-defined nature and possibility of getting an efficient estimation.
Being a member of the exponential family, the belief distributions $b_{\bm\eta}(\theta)$ construct a statistical manifold with local co-ordinates $\bm\eta$~\cite{amari2007infogeo}.
Theorem~\ref{thm:unique} and~\ref{thm:clt} validate these claims in case of BelMan.

\textbf{Bernoulli Bandits.}\label{app:belman-exponential-family}
In the case of Bernoulli bandits, we assume that drawing an arm returns the rewards $1$ and $0$ with probability $\theta$ and $1-\theta$ respectively.
Thus, the reward distribution of the $a^{\mathrm{th}}$ arm is $f_{\theta_a}(X) \triangleq \mathrm{Ber}(\theta_a)$.
Following the Bayesian approach, we choose the conjugate prior to begin with.
Thus, we keep the prior belief over each arm as a beta distribution with shape parameters $\lbrace \alpha^a \rbrace_{a=1}^K$ and $\lbrace \beta^a \rbrace_{a=1}^K$.
After $t$-iterations the prior over the probability of success of the $a^{\mathrm{th}}$ arm is
\begin{align*}
b^a_t (\theta_a) \triangleq \mathrm{Beta}(\theta_a; \alpha^a_t , \beta^a_t) = \frac{1}{B(\alpha^a_t , \beta^a_t)} \theta_a^{\alpha^a_t -1} (1- \theta_a)^{\beta^a_t - 1},
\end{align*}
for $\alpha^a_t , \beta^a_t > 0$ and $\theta_a \in (0,1)$.
Here, $\alpha^a_t$ and $\beta^a_t$ are the number of successes and failures, respectively, for the arm $a$ till iteration $t$.
We begin with both $\alpha^a_0$ and $\beta^a_0$ to be 1 for all arms.
This amounts to the uniform distribution over 0 and 1.
This initialisation allows us to choose all the arms with equal probability and without any initial bias.
We update this belief eventually as we further draw the arms and compute it using BelMan.
Under this specific setting of beta prior and Bernoulli reward, we compute the targeted KL-divergence of BelMan as
\begin{align*}
&\sum_{a=1}^K D_{\mathrm{KL}}\left(\mathbb{P}^a_t(X,\theta)\|\bar{\mathbb{Q}}_{t-1}(X,\theta)\right)\\
&= \sum_{a=1}^K [-\frac{1}{\tau(t)}\frac{\alpha^a_t}{N^a_t} - \log\left(B\left({\alpha}^a_t,{\beta}^a_t\right)\right) + (\alpha^a_t - \bar{\alpha}_{t-1})\Psi(\alpha^a_t) + (\beta^a_t - \bar{\beta}_{t-1})\Psi(\beta^a_t)-\\ &(N^a_t - \bar{N}_{t-1})\Psi(N^a_t) ] + K\log\left(\frac{\bar{\alpha}_{t-1} \exp(\frac{1}{\tau(t)}) + \bar{\beta}_{t-1}}{\bar{N}_{t-1}}\right) + K\log\left(B\left(\bar{\alpha}_{t-1},\bar{\beta}_{t-1}\right)\right).
\end{align*}
Here, $N^a_t= \alpha^a_t + \beta^a_t$ is the total number of times the $jth$ arm is played till the $nth$ iteration, $\bar{N} = \bar\alpha + \bar\beta$ and $\Psi$ is the digamma function~\cite{bernardo1976digamma} defined as the derivative of the logarithm of gamma function, i.e. $\frac{d}{da}\left(\log \Gamma(a)\right)$.

In Line 4 of Algorithm~\ref{alg:belman}, we first perform the I-projection to decide which arm $a_t$ to draw to minimize the KL-divergence.
Following this, we update the pseudobelief using I-projection in Line 9 of Algorithm~\ref{alg:belman}.
In order to perform this update, we find out such $\bar{\alpha}$ and
  $\bar{\beta}$ that minimize the objective and update the pseudobelief accordingly.
The presence of pseudobelief offers BelMan a chance to explore the less successful arms to minimize the entropy, while the Focal distribution creates the scope of exploiting the present information of the best arm.

\textbf{Exponential Bandits.}
The \emph{exponential distribution} is another member of the exponential family.
For a given positive \emph{rate parameter} $\theta_a$, the reward distribution of arm $a$ of exponential bandit is $f_{\theta_a}(X) \triangleq \theta_a \exp(-\theta_a X)$ for $X \in [0, \infty)$.
Following the structure of Sections~\ref{sec:expfamily} and the previous
  Bernoulli case, we obtain the gamma distribution, another member of the exponential family, as the conjugate prior.
After the $t^{\mathrm{th}}$ iteration, the belief distribution corresponding to $a^{\mathrm{th}}$ arm is expressed as
\begin{align*}
b^a_t (\theta_a) \triangleq \mathrm{Gamma}(\theta_a; \alpha^a_t , \beta^a_t) = \frac{{\beta^a_t}^{\alpha^a_t}}{\Gamma(\alpha^a_t)} {\theta_a}^{\alpha^a_t -1} \exp(- \theta_a\beta^a_t),
\end{align*}
for both shape and rate parameters $\alpha^a_t , \beta^a_t > 0$.
Here, $\alpha^a_t$ and $\beta^a_t$ are, respectively, the number of times the arm $a$ is played and sum of the rewards obtained by playing the arm till iteration $t$.
As we update using Equation~\eqref{eqn:beliefupdate}, we get gamma distributions with parameters $\alpha^a_{t+1} = \alpha^a_t +1$, and $\beta^a_{t+1} = \beta^a_t +x_t$ if the arm $a$ is played and a reward $x_t$ is obtained.
Under this specific setting of gamma prior and exponential reward, we compute the targeted KL-divergence of BelMan as
\begin{align*}
&\sum_{a=1}^K D_{\mathrm{KL}}\left(\mathbb{P}^a_t(X,\theta)\|\bar{\mathbb{Q}}(X,\theta)\right)\\ &= \sum_{a=1}^K [ -\frac{1}{\tau(t)}\frac{\alpha^a_t}{\beta^a_t} - \log\left(\Gamma\left({\alpha}^a_t\right)\right) + (\alpha^a_t - \bar{\alpha}_{t-1})\Psi(\alpha^a_t) - \frac{\alpha^a_t}{\beta^a_t}(\beta^a_t - \bar{\beta}_{t-1})\\ &+ \bar{\alpha}_{t-1} \log\beta^a_t] + K\log \bar{Z}_t + K\log\left(\Gamma\left(\bar{\alpha}_{t-1}\right)\right) - K\bar{\alpha}_{t-1}\log\bar{\beta}_{t-1}.
\end{align*}
We incorporate this analytical form in Algorithm~\ref{alg:belman} and
  update it as mentioned in the Bernoulli case.
\end{toappendix}
\section{Empirical Performance Analysis}\label{sec:parametric}
\paragraph{Exploration--exploitation bandit problem.}
We evaluate the performance of BelMan for two exponential family distributions -- Bernoulli and exponential.
They stand for discrete and continuous rewards respectively.
We use the pymaBandits library~\cite{pymabandits} for implementation of
all the algorithms except ours, and run it on MATLAB 2014a.
We plot the evolution of the mean and the 75 percentile of cumulative regret and number of suboptimal draws.
For each instance, we run experiments for 25 runs each consisting of 1000 iterations.
We begin with uniform distribution over corresponding parameters as the initial prior distribution for all the Bayesian algorithms.

\begin{toappendix}
  \label{app:experiments}
\begin{figure*}[p]
		\vspace*{-1.5cm}
		\hspace*{-1.4cm}
		\includegraphics[scale=0.35,trim={5cm 12cm 5cm
                1cm},clip]{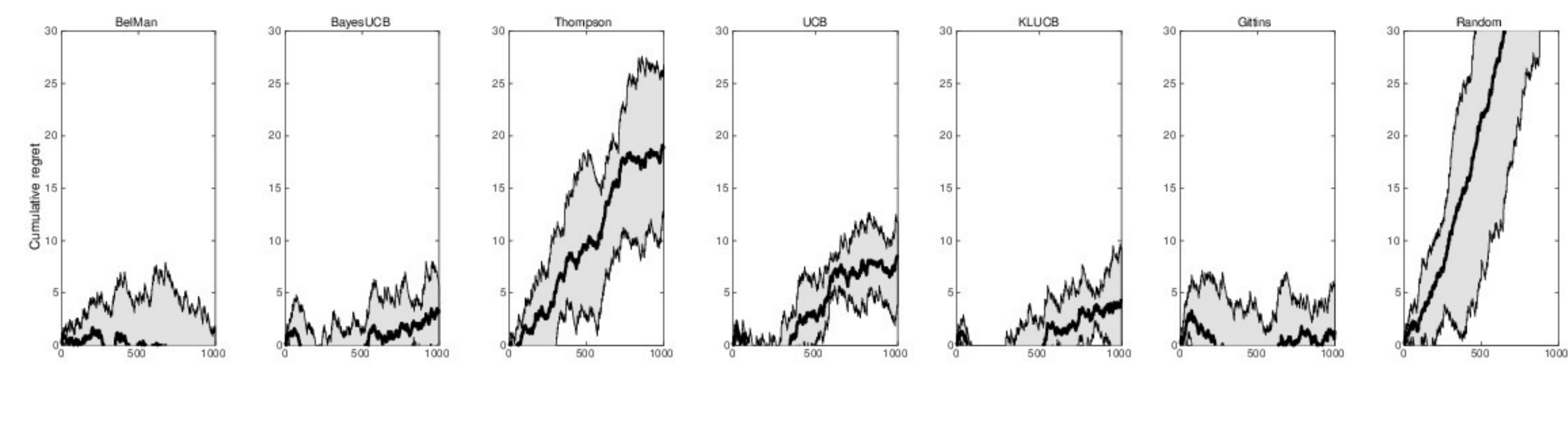}\vspace*{-.9cm}
		\caption{Evolution of cumulative regret (top), and number of suboptimal draws (bottom) for 2-arm Bernoulli bandit with expected rewards 0.8 and 0.9 for 1000 iterations. The dark black line shows the average over 25 runs. The grey area shows the 75 percentile.}\label{fig:ber_2_80_90_reg}
		\hspace*{-1.4cm}
		\includegraphics[scale=0.35,trim={5cm 12cm 5cm
                1cm},clip]{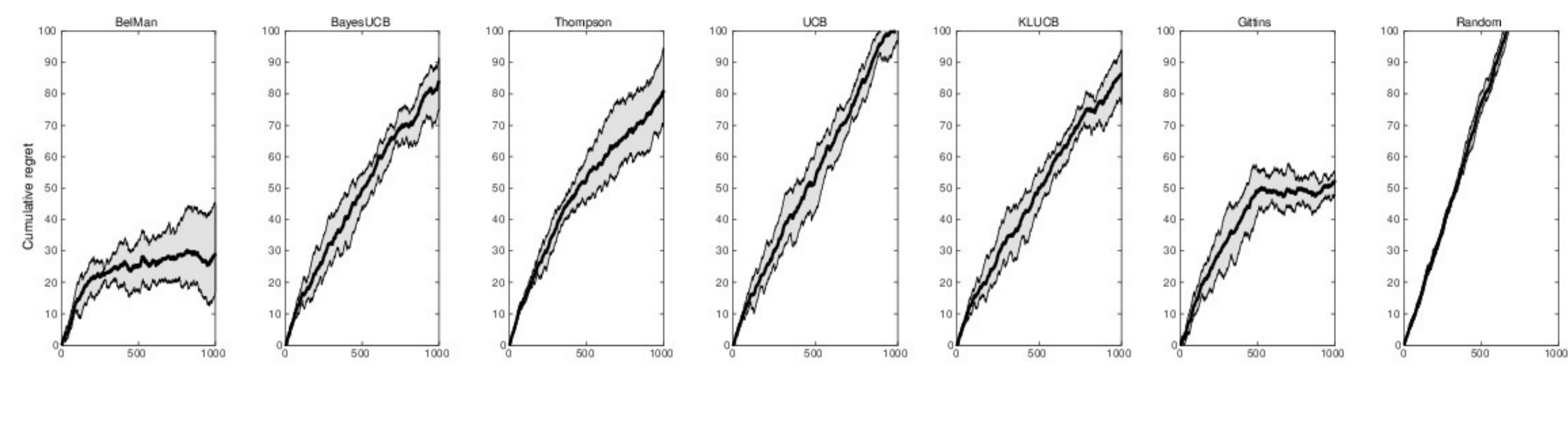}\vspace*{-.9cm}
		\caption{Evolution of cumulative regret (top), and number of suboptimal draws (bottom) for 20-arm Bernoulli bandit with expected rewards [0.25 0.22 0.2 0.17 0.17 0.2 0.13 0.13 0.1 0.07 0.07 0.05 0.05 0.05 0.02 0.02 0.02 0.01 0.01 0.01] for 1000 iterations.}\label{fig:ber_20_reg}
		\hspace*{-1.4cm}
		\includegraphics[scale=0.35,trim={5cm 13cm 4cm 1cm},clip]{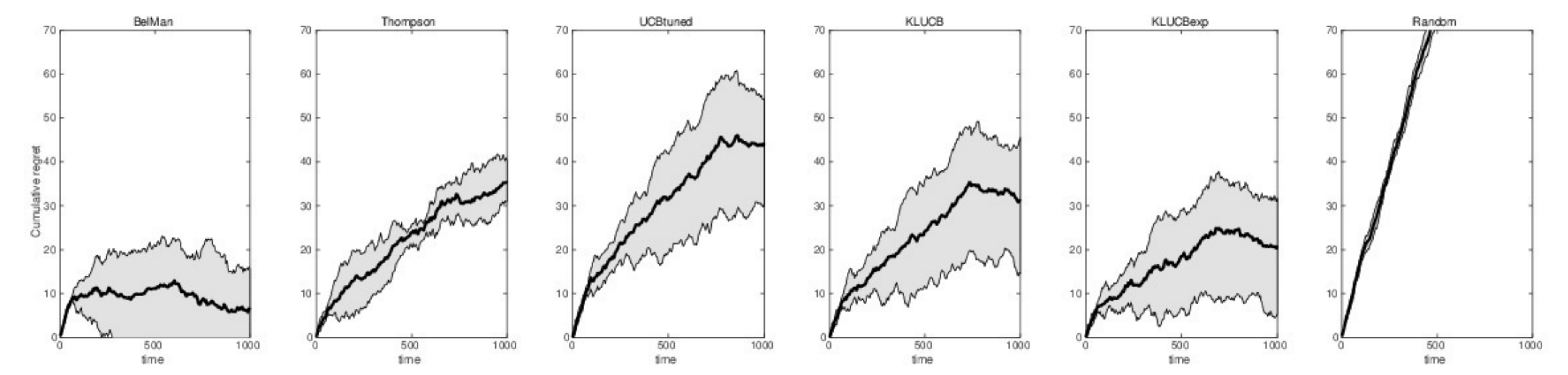}\vspace*{-.8cm}
		\caption{Evolution of cumulative regret (top), and number of suboptimal draws (bottom) for 5-arm bounded exponential bandit with expected rewards 0.2, 0.25, 0.33, 0.5, and 1.0 for 1000 iterations.}\label{fig:exp_5_reg}
		\hspace*{-3.1cm}
		\includegraphics[scale=0.42]{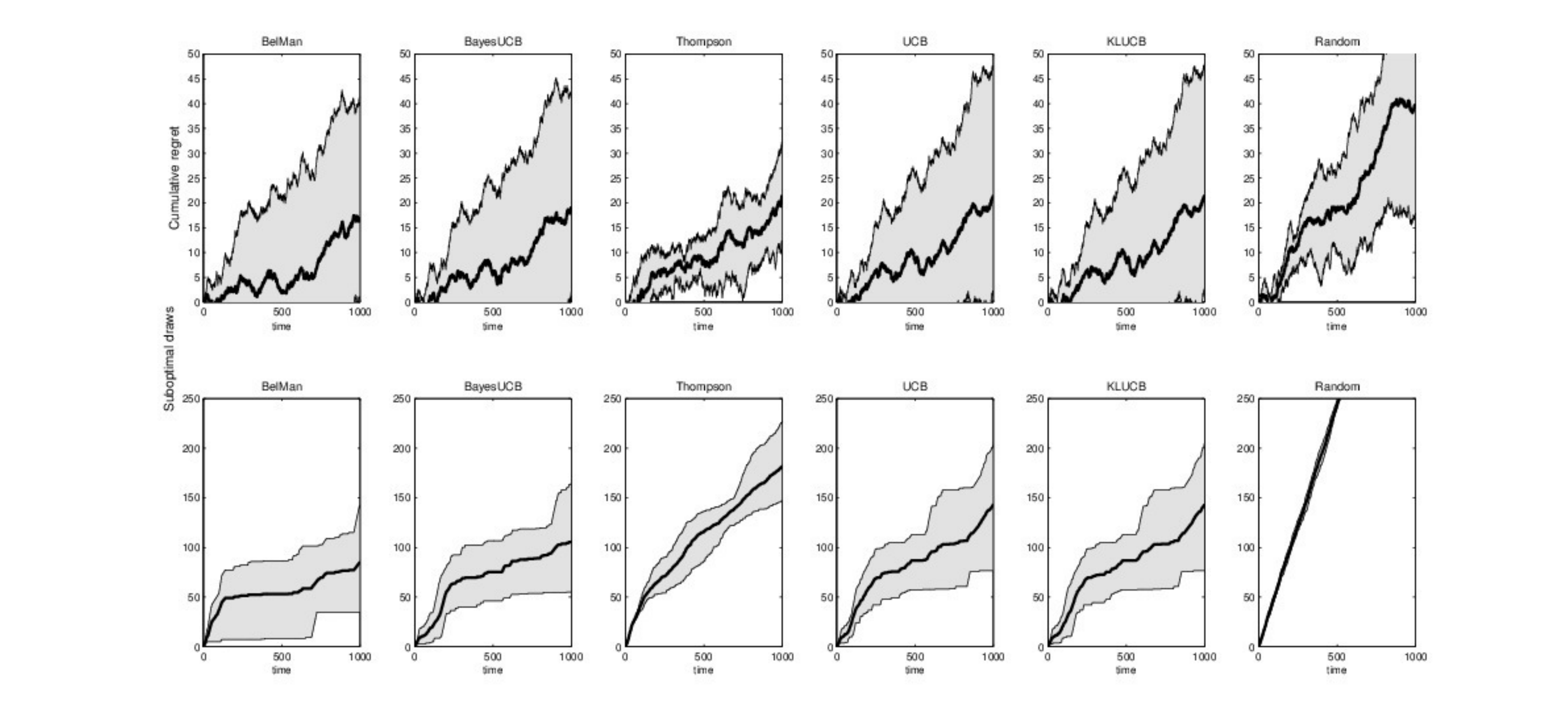}\vspace*{-1.2cm}
		\caption{Evolution of cumulative regret (top), and number of suboptimal draws (bottom) for 500 iterations for 2-arm Bernoulli bandit with means $0.45$ and $0.55$.}~\label{fig:ber_2_45_55}
\end{figure*}

Figure~\ref{fig:ber_2_80_90_reg},~\ref{fig:ber_20_reg}, and~\ref{fig:exp_5_reg} show the evolution of cumulative regret with number of iterations for the three cases whose number of suboptimal arm draws are reported in Figure~\ref{fig:ber_2_80_90},~\ref{fig:ber_20}, and~\ref{fig:exp_5}, respectively.

We also experimented on another 2-arm bandit scenario with means 0.45 and 0.55.
Figures~\ref{fig:ber_2_45_55} depicts the evolution of cumulative regret and suboptimal draws for BelMan and the other competing algorithms.
Similar to Figure~\ref{fig:ber_2_45_55}, we observe the cumulative regret of BelMan grows at first linearly and then it transits to a state of slow growth.
Except showing this ideal behaviour, BelMan performs competitively with the contending algorithms.
This shows its efficiency as a candidate solution to the exploration--exploitation bandit.

Figure~\ref{fig:ber_10} shows performance for 10-arm Bernoulli bandit.
For this setup, BelMan outperforms other algorithms.
We also observe though the number of arms increases from Figure~\ref{fig:ber_2_45_55} to Figure~\ref{fig:ber_10} that performance of all algorithms is comparatively better in the first case.
This is explainable from the fact that hardness of minimising cumulative regret increases as the number of arms increases.
Beside that, as more arms with identical or almost identical distributions appear, the algorithm requires more exploration to separate them and to determine which one is optimal.
The difference in performance between Figure~\ref{fig:ber_2_45_55} and~\ref{fig:ber_2_80_90} indicates this.
\begin{figure*}[p]
\vspace*{-1.5cm}
\hspace*{-1.5cm}
\includegraphics[scale=0.35]{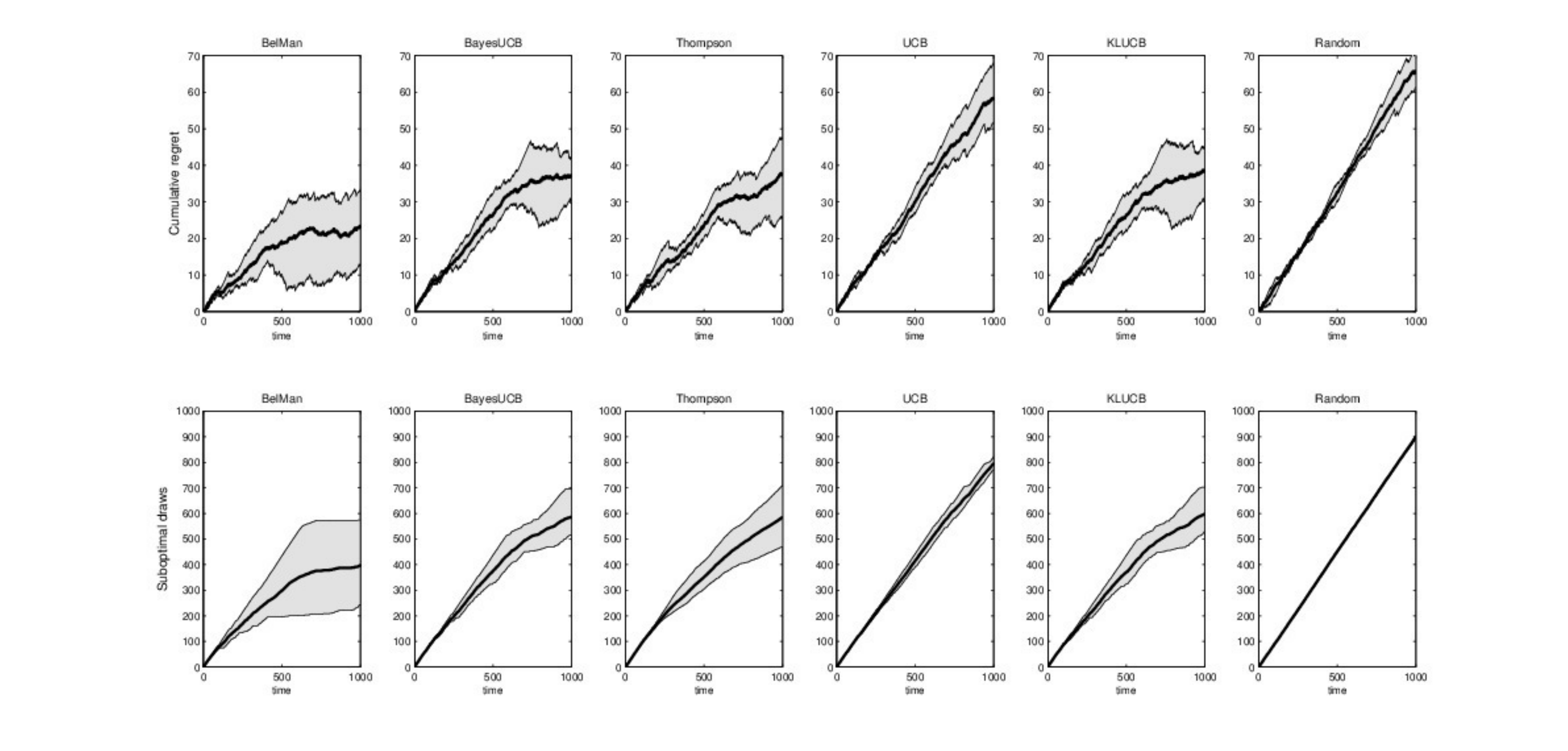}\vspace*{-1.0cm}
\caption{Evolution of cumulative regret (top), and number of suboptimal draws (bottom) for 500 iterations for 10-arm Bernoulli bandit with means $\lbrace 0.1, 0.05, 0.05, 0.05, 0.02, 0.02, 0.02, 0.01, 0.01, 0.01\rbrace$. The dark black line shows the average. The grey area shows 75 percentile.}~\label{fig:ber_10}
\hspace*{-1.5cm}
\includegraphics[scale=0.35]{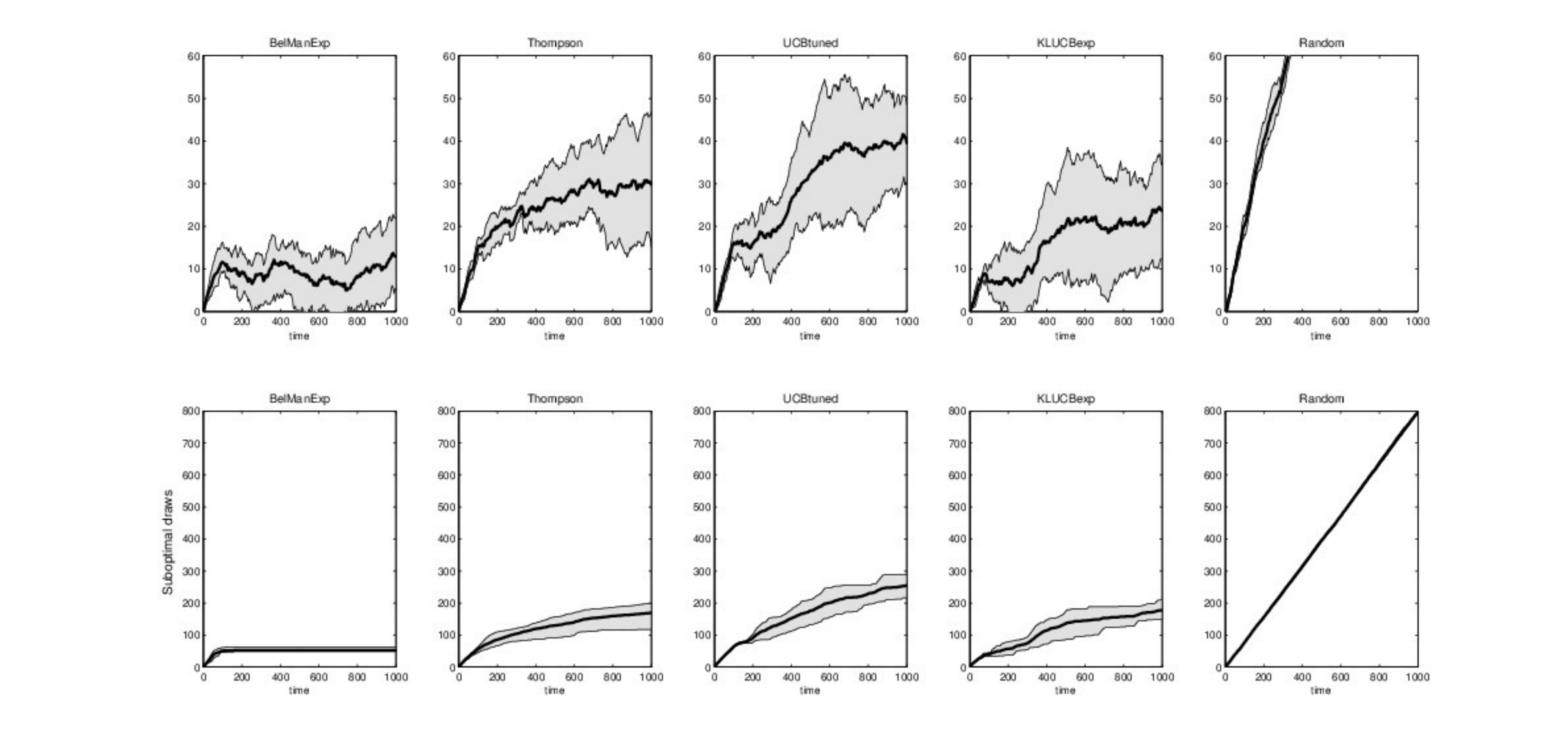}\vspace*{-1.0cm}
\caption{Evolution of cumulative regret (top), and number of suboptimal draws  (bottom) for 1000 iterations for 5-arm unbounded exponential bandit with parameters $\lbrace 0.2, 0.25, 0.33, 0.5, 1.0\rbrace$.}~\label{fig:exp_5_leq1}
\hspace*{-1.5cm}
\includegraphics[scale=0.35]{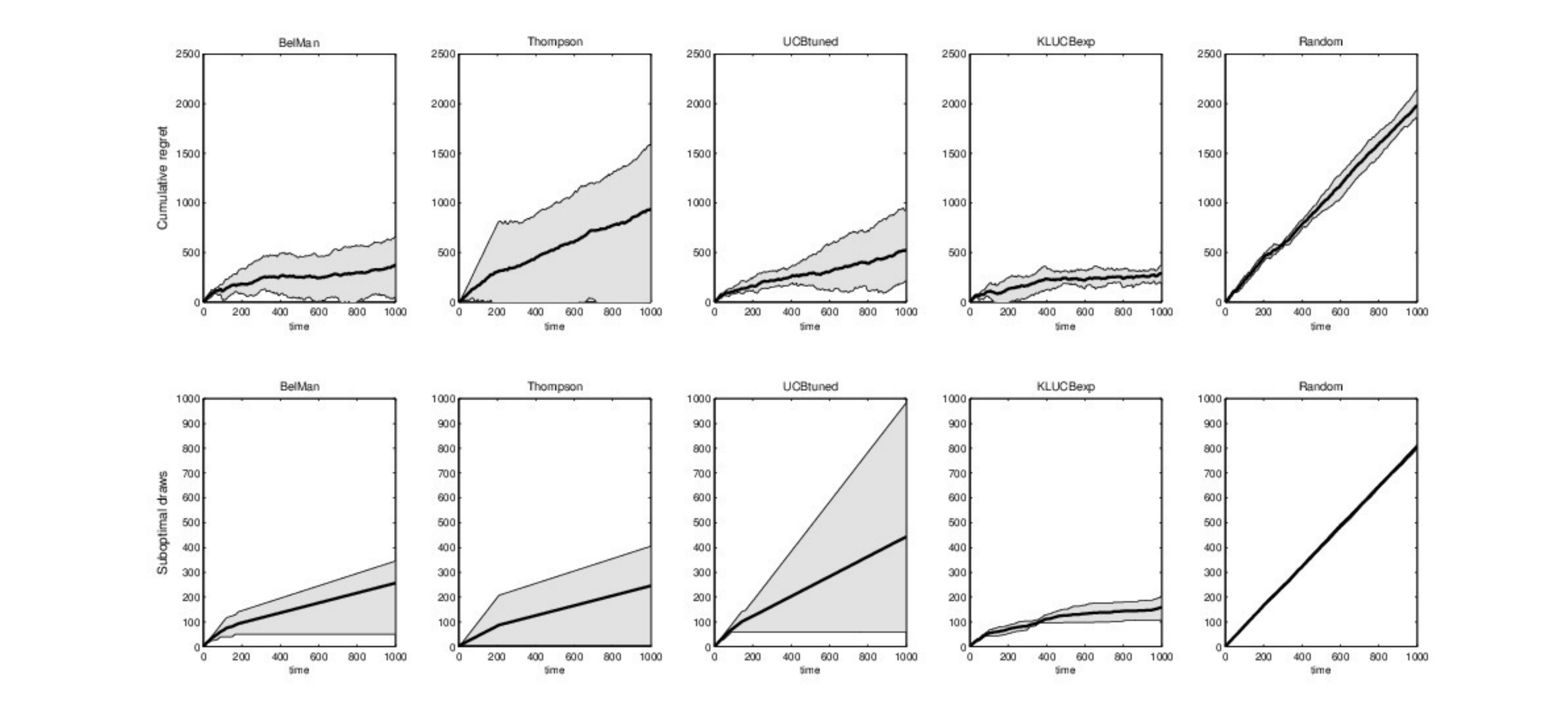}\vspace*{-1.0cm}
\caption{Evolution of cumulative regret (top), and number of suboptimal draws  (bottom) for 1000 iterations for 5-arm unbounded exponential bandit with parameters $\lbrace 1, 2, 3, 4, 5\rbrace$.}~\label{fig:exp_5_geq1}
\end{figure*}

We finally tested BelMan on an exponential bandit consisting of 5-arms with expected rewards $\lbrace 0.2, 0.25,$ $ 0.33, 0.5, 1.0\rbrace$.
We compare performance of BelMan with state-of-the-art frequentist method tailored for exponential distribution of rewards, called KL-UCBExp~\cite{garivier2011klucb}.
We also compare it with Thompson sampling, UCBtuned and uniform sampling method (Random).
The results are shown in Figure~\ref{fig:exp_5_leq1} and~\ref{fig:exp_5_geq1}.
Since the formulation is oblivious to boundedness of the distribution, we choose to validate also on unbounded rewards.
In Figure~\ref{fig:exp_5_leq1}, it outperforms all the other algorithms.
In Figure~\ref{fig:exp_5_geq1}, though KL-UCBexp performs the best, performance of BelMan is still competitive with it.

These results validate BelMan's claim as a generic solution to a wide range of bandit problems.
\end{toappendix}
We compare the performance of BelMan with frequentist
methods like UCB~\cite{auer2002finite} and
KL-UCB~\cite{garivier2011klucb}, and Bayesian methods like Thompson
sampling~\cite{thompson1933} and Bayes-UCB~\cite{kaufmann2012bayesucb}.
For Bernoulli bandits, we also compare with Gittins index~\cite{gittins1979} which is the optimal algorithm for Markovian finite arm independent bandits with discounted rewards. Though we are not specifically interested in the discounted case, Gittins' algorithm is indeed transferable to the finite horizon setting with slight manipulation. Though it is often computationally intractable, we use it as the optimal baseline for Bernoulli bandits.
We also plot performance of the uniform sampling method (\emph{Random}), as a na{\"i}ve baseline.

From Figures~\ref{fig:ber_2_80_90}, \ref{fig:ber_20}, and \ref{fig:exp_5}, we observe that at the very beginning the number of suboptimal draws of BelMan grows linearly and then transitions to a state of slow growth.
This initial linear growth of suboptimal draws followed by a logarithmic
growth is an intended property of any optimal bandit algorithm as can be
seen in the performance of competing algorithms and also pointed out by~\cite{garivier2016explore}:
an initial phase dominated by exploration and a second phase dominated by exploitation.
The phase change indicates the ability of the algorithm to reduce uncertainty by learning after a certain number of iterations, and to find a trade-off between exploration and exploitation.
For the 2-arm Bernoulli bandit ($\theta_1=0.8, \theta_2=0.9$), BelMan performs comparatively well with respect to the contending
algorithms, achieving the phase of exploitation faster than others, with significantly less variance.
Figure~\ref{fig:ber_20} depicts similar features of BelMan for 20-arm Bernoulli bandits (with means 0.25, 0.22, 0.2, 0.17, 0.17, 0.2, 0.13, 0.13, 0.1, 0.07, 0.07, 0.05, 0.05, 0.05, 0.02, 0.02, 0.02, 0.01, 0.01, and 0.01).
Since more arms ask for more exploration and more suboptimal draws, all algorithms show higher regret values.
On all experiments performed, 
BelMan outperforms the competing approaches.
\begin{figure*}[t!]
	\vspace*{-2em}
	\begin{minipage}{0.48\textwidth}
	\hspace{-2em}\includegraphics[scale=0.2,trim={4cm 4cm 1cm 1cm}, clip]{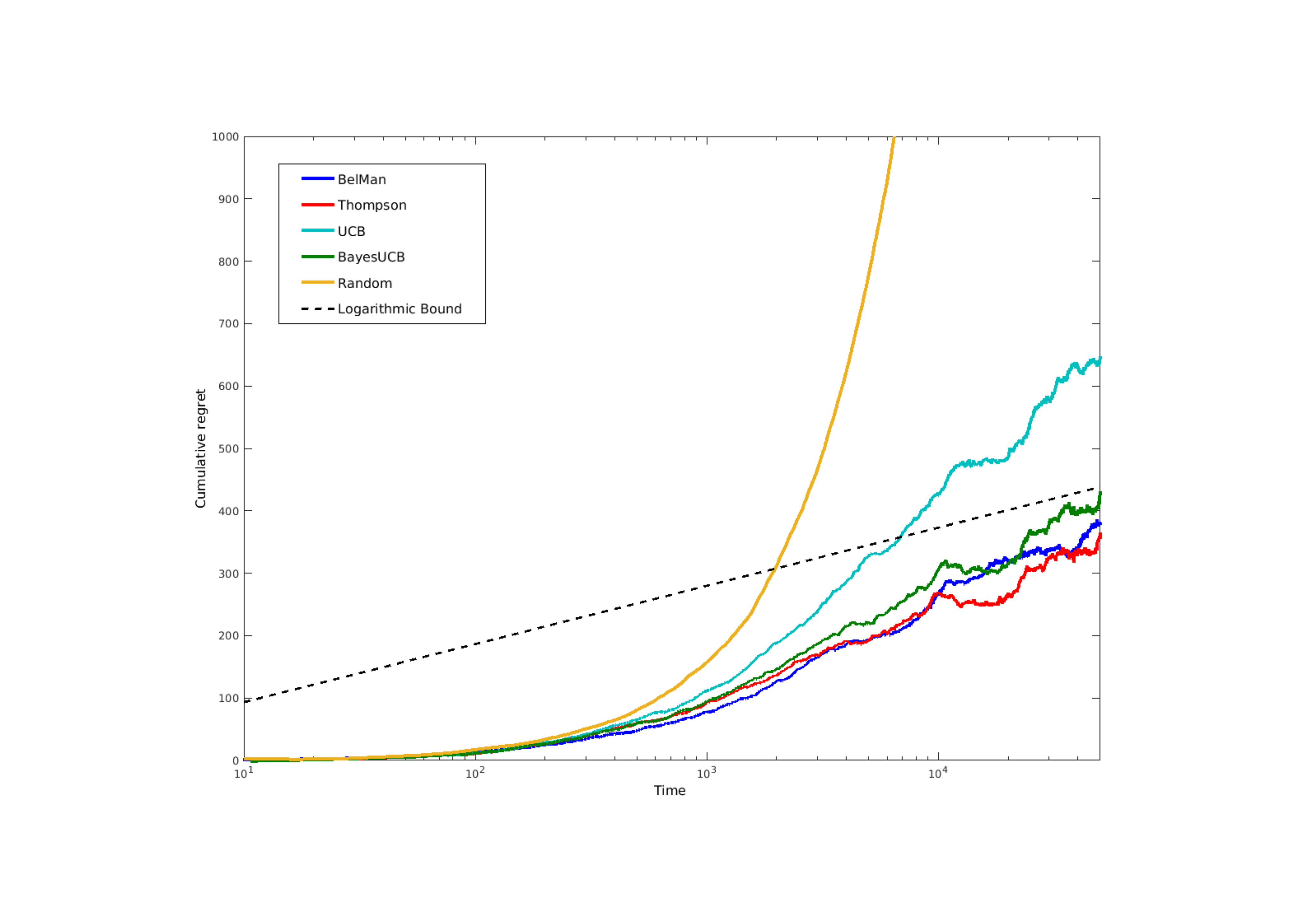}
		\vspace*{-0.5cm}
		\caption{Evolution of (mean) regret for exploration--exploitation 20-arm Bernoulli bandit setting of Figure~\ref{fig:ber_20} with horizon=50,000.}\vspace*{-0.3cm}\label{fig:ber_long_horizon}
	\end{minipage}
	\hfill
	\begin{minipage}{0.48\textwidth}
		\hspace{-2em}\includegraphics[scale=0.2,trim={4cm 4cm 1cm 1cm}, clip]{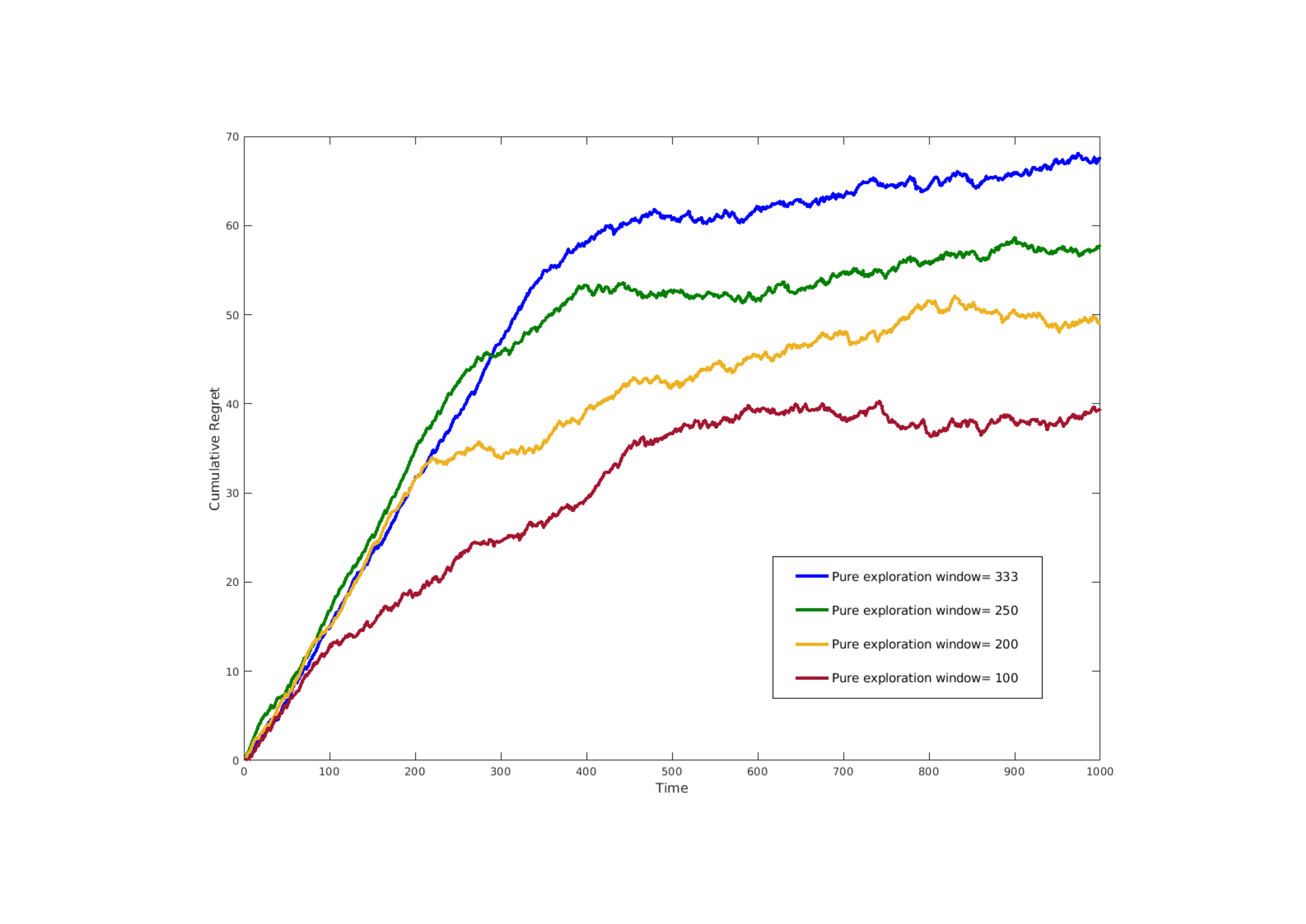}
		\caption{Evolution of (mean) cumulative regret for two-phase 20-arm Bernoulli bandits.}\vspace*{-0.3cm}
		\label{fig:ber_2phase_windows}
	\end{minipage}\vspace*{.9em}
\end{figure*}
%
%
%
%
We also simulated BelMan on exponential bandits: 5 arms with expected rewards $\lbrace 0.2, 0.25, 0.33, 0.5, 1.0\rbrace$.
Figure~\ref{fig:exp_5} shows that BelMan performs more efficiently than state-of-the-art methods for exponential reward distributions- Thompson
  sampling, UCBtuned~\cite{auer2002finite}, KL-UCB, and KL-UCB-exp, a method tailored for exponential distribution of rewards~\cite{garivier2011klucb}.
This demonstrates BelMan's broad applicability and efficient performance in complex scenarios.

We have also run the experiments 50 times with horizon 50\,000 for the 20 arm Bernoulli bandit setting of Figure~\ref{fig:ber_20} to verify the asymptotic behaviour of BelMan. Figure~\ref{fig:ber_long_horizon} shows that BelMan's regret gradually becomes linear with respect to the logarithmic axis.  Figure~\ref{fig:ber_long_horizon} empirically validates BelMan to achieve logarithmic regret like the competitors which are theoretically proven to reach logarithmic regret.

\textbf{Two-phase reinforcement learning problem.} In this experiment, we simulate a two-phase
setup, as in~\cite{putta2017twophaseepisodic}: the agent first does pure
exploration for a fixed number of iterations, then move to 
exploration--exploitation. This is possible since BelMan supports both modes and can transparently switch.
The setting is that of the 20-arm Bernoulli bandit in Figure~\ref{fig:ber_20}.
The two-phase algorithm is exactly BelMan (Algorithm~\ref{alg:belman}) with $\tau(t) = \infty$ for an initial phase of length $T_{\mathrm{EXP}}$ followed by the decreasing function of $t$ as indicated previously.
Thus, BelMan gives us a single algorithmic framework for three setups of bandit  problems-- pure exploration, exploration--exploitation, and two-phase learning. We only have to choose a different $\tau(t)$ depending on the problem addressed. This supports BelMan's claim as a generalised, unified framework for stochastic bandit problems.

We observe a sharp phase transition in Figure~\ref{fig:ber_2phase_windows}.
While the pure exploration version acts in the designated window length, it explores almost uniformly to gain more information about the reward distributions.
We know for such pure exploration the cumulative regret grows linearly with iterations.
Following this, the growth of cumulative regret decreases and becomes sublinear.
If we also compare it with the initial growth in cumulative regret and
suboptimal draws of BelMan in Figure~\ref{fig:ber_20}, we observe that
the regret for the exploration--exploitation phase is less than
that of regular BelMan exploration--exploitation.
Also, with increase in the window length the phase transition becomes sharper as the growth in regret becomes very small.
In brief, there are three major lessons of this experiment.
First, Bayesian methods provide an inherent advantage in leveraging
prior knowledge (here, accumulated in the first phase).
Second, a pure exploration phase helps in improving the performance
during the exploration--exploitation phase.
Third, we can leverage the exposure to control the exploration--exploitation trade-off.

\begin{figure*}[t!]%
	\vspace*{-1.5em}
	\centering
	\subfigure[][Q-ThS]{%
		\label{fig:ex3-35-a}%
		\includegraphics[width=0.46\textwidth]{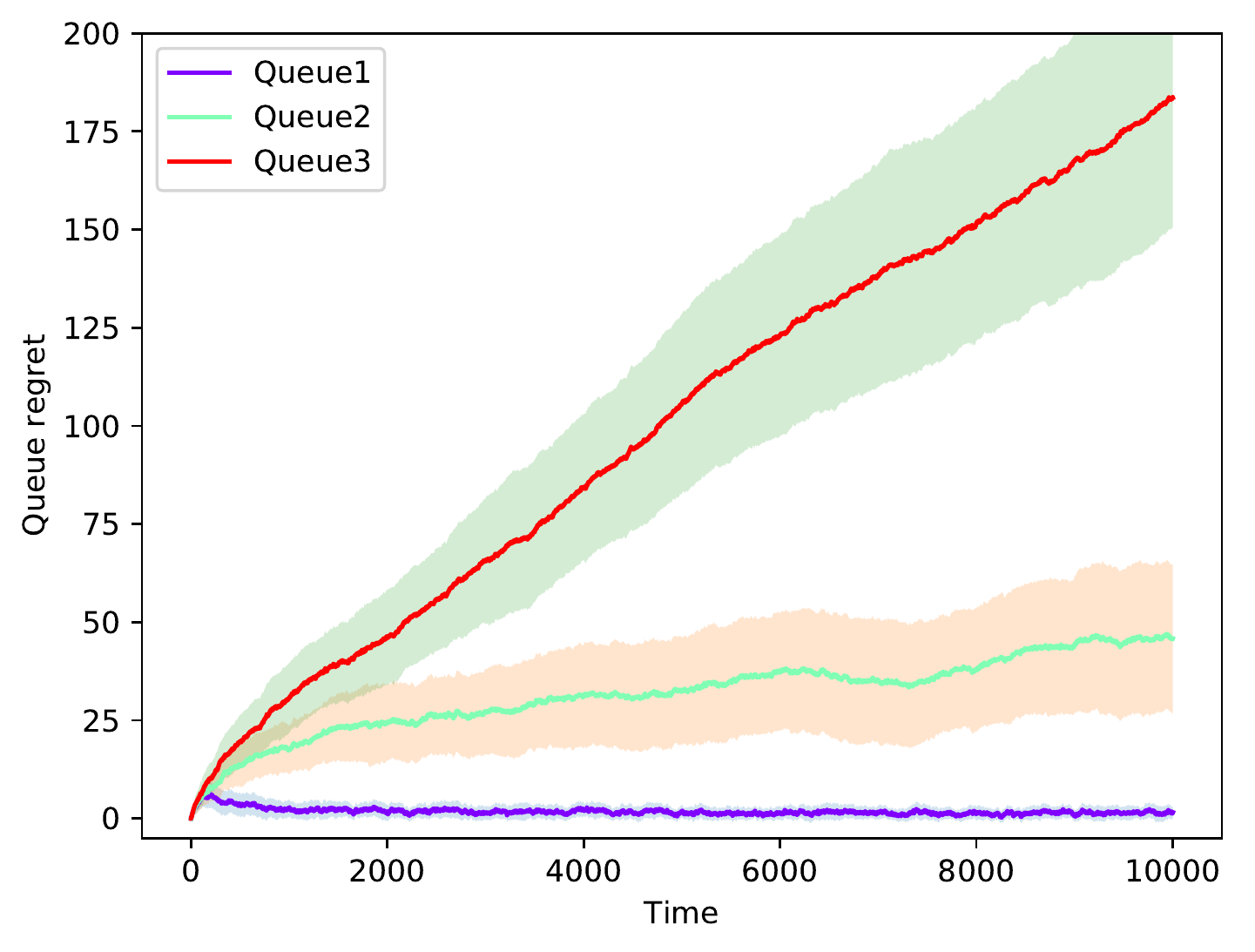}}%
	\hspace{8pt}%
	\subfigure[][Q-UCB]{%
		\label{fig:ex3-35-b}%
		\includegraphics[width=0.46\textwidth]{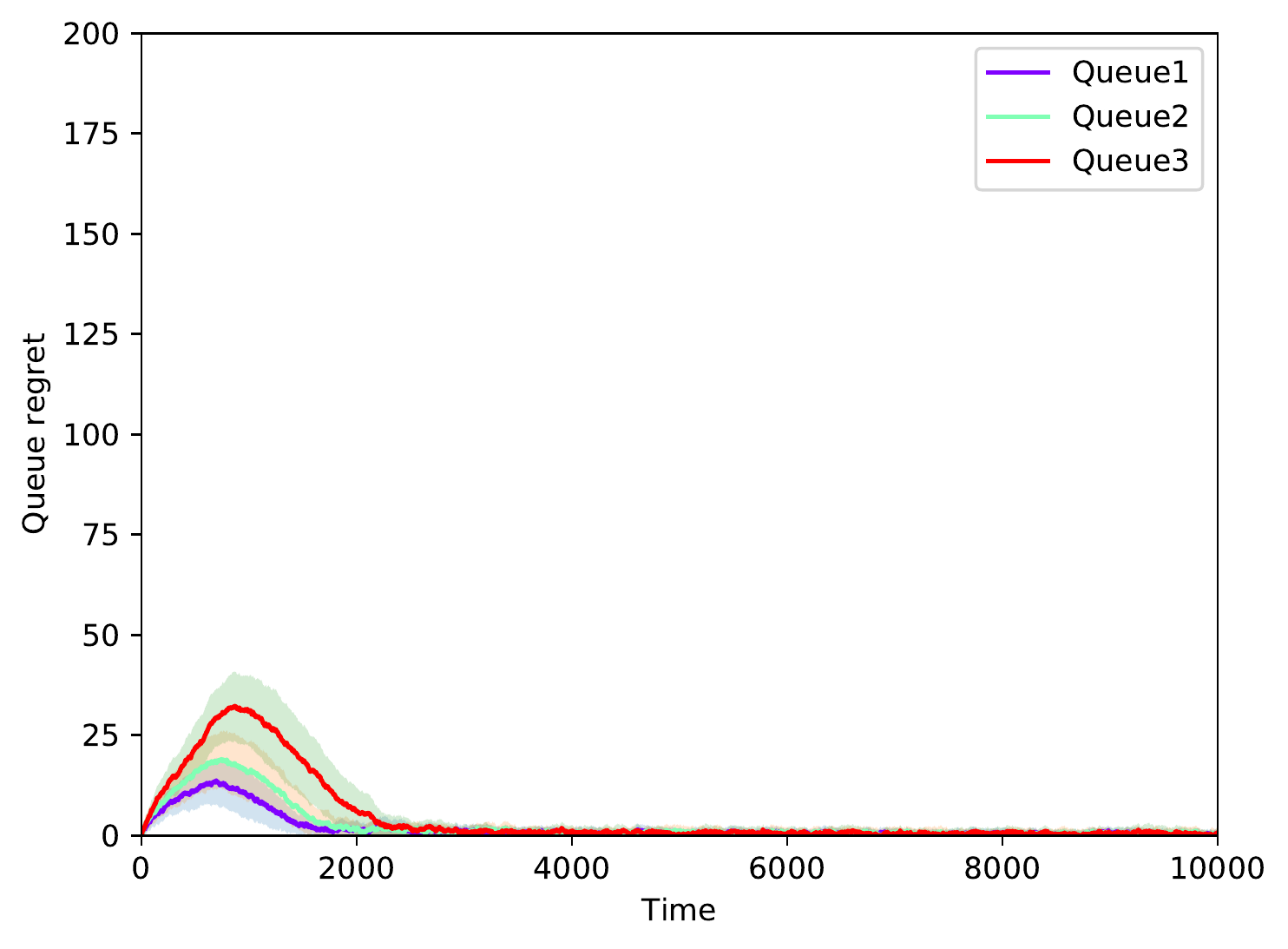}}\vspace*{-1em}
	\subfigure[][Thompson sampling]{%
		\label{fig:ex3-35-c}%
		\includegraphics[width=0.46\textwidth]{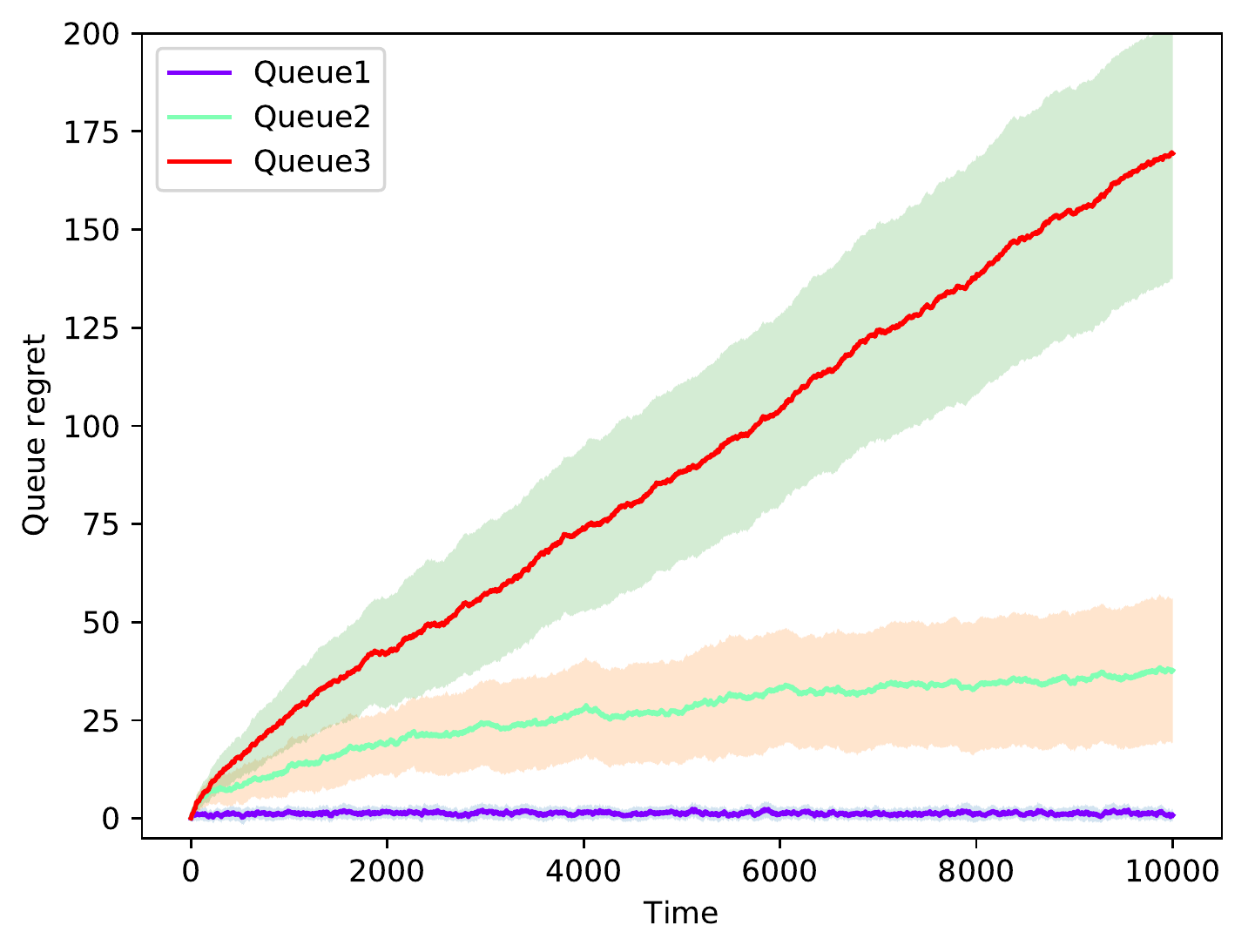}}%
	\hspace{8pt}%
	\subfigure[][BelMan]{%
		\label{fig:ex3-35-d}%
		\includegraphics[width=0.46\textwidth]{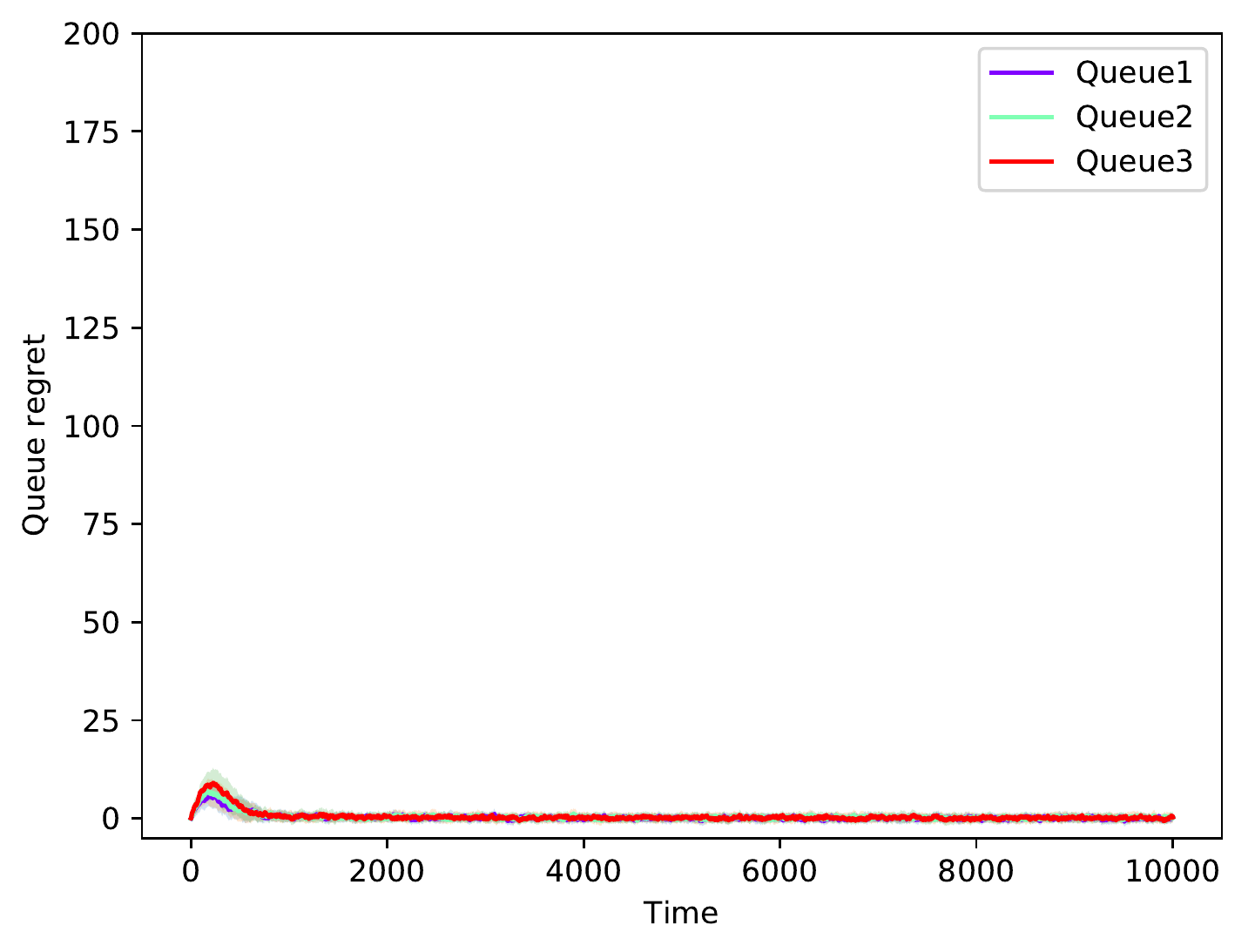}}
		\vspace*{-1.2em}
	\caption{Queue regret for single queue and 5 server setting with Poisson arrival with arrival rate 0.35 and Bernoulli service distribution with service rates [0.5,0.33,0.33,0.33,0.25], [0.33,0.5,0.25,0.33,0.25], and [0.25,0.33,0.5,0.25,0.25] respectively. Each experiment is performed 50 times for a horizon of 10,000.}\label{fig:ex4}%
\end{figure*}
\section{Application to Queueing Bandits}\label{sec:queueing}
We instantiate BelMan for the problem of scheduling jobs in a
multiple-server multiple-queue system with known arrival rates and unknown service rates. The goal of the agent is to choose such a server for the given system such that the total queue length, i.e. the jobs waiting in the queue, will be as less as possible. This problem is referred as the queueing bandit~\cite{krishnasamy2016regret}.

We consider a discrete-time queueing system with $1$ queue and $K$ servers. The servers are indexed by $a \in \lbrace 1, \ldots, K \rbrace$. Arrivals to the queue and service offered by the servers are assumed to be independent and identically distributed across time.
The mean arrival rate is $\lambda \in \mathbb{R}^+$.
The mean service rates are denoted by $\bm{\mu} \in  \lbrace \mu_{a} \rbrace_{a=1}^{K}$, where $\mu_{a}$ is the service rate of server $a$.
At a time, a server can serve the jobs coming from a queue only.
We assume the queue to be stable i.e, $\lambda < \max\limits_{a\in[K]} \mu_{a}$. 
Now, the problem is to choose a server at each time $t \in [T]$ such that the number of jobs waiting in queues is as less as possible.
The number of jobs waiting in queues is called the \emph{queue length} of the system.
If the number of arrivals to the queues at time $t$ is $A(t)$ and $S(t)$ is the number of jobs served, the queue length at time $t$ is defined as \(Q(t) \triangleq Q(t-1) + A(t) - S(t), \)
where $Q: [T] \rightarrow \mathbb{R}^{\geq 0}$, $A: [T] \rightarrow \mathbb{R}^{\geq 0}$, and $S: [T] \rightarrow \mathbb{R}^{\geq 0}$.
The agent, which is the scheduling algorithm in this case, tries to minimise this queue length for a given horizon $T > 0$.
The arrival rates are known to the scheduling algorithm but the service rates are unknown to it.
This create the need to learn about the service distributions, and in turn, engenders the exploration-exploitation dilemma.

Following the bandit literature, \cite{krishnasamy2016regret} proposed to use \emph{queue regret} as the performance measure of a queueing bandit algorithm.
Queue regret is defined as the difference in the queue length if a bandit algorithm is used instead of an optimal algorithm with full information about the arrival and service rates.
Thus, the \emph{optimal algorithm} $\mathrm{OPT}$ knows all the arrival and service rates, and allocates the queue to servers with the best service rate.
Hence, we define the queue regret of a queueing bandit algorithm
\(	\Psi(t) \triangleq \mathbb{E} \left[Q(t) - Q^{\mathrm{OPT}}(t)\right].
\)
In order to keep the bandit structure, we assume that both the queue length $Q(t)$ of algorithm $\mathcal{A}$ and that of the optimal algorithm $Q^{\mathrm{OPT}}(t)$ starts with the same stationary state distribution $\nu(\lambda, \bm{\mu})$.

We show experimental results for the $M/B/K$ queueing bandits. We assume the arrival process to be Markovian, and the service process to be Bernoulli. 
The arrival process being Markovian implies that the stochastic process describing the number of arrivals is therefore $A\left(t\right)$ have increments independent of time.
This makes the distribution of $A(t)$ to be a Poisson distribution~\cite{durrett2010probability} with mean arrival rate $\lambda$. 
We denote $B_{a}(\mu_a)$ is the Bernoulli distribution of the service time of server $a$.
It implies that the server processes a job with probability $\mu_a \in (0,1)$ and  refuses to serve it with probability $1 - \mu_a$.
The goal is to perform the scheduling in such a way that the queue regret will be minimised.
The experimental results in Figure~\ref{fig:ex4} depict that BelMan is more stable and efficient than the competing algorithms: Q-UCB, Q-Thompson sampling, and Thompson sampling.
We observe that in queues 2 and 3 the average service rates are lower than the corresponding arrival rates.
Due to this inherent constraint, the queue 2 and 3 can have unstable queueing systems if the initial exploration of the algorithm does not damp fast enough. Though the randomisation of Thompson sampling is good for exploration but in this case playing the suboptimal servers can induce instability which affects the total performance in future.

\section{Related Work}

\cite{bellman1956} posed the problem of discounted reward bandits with infinite horizon as a single-state Markov decision process~\cite{gittins1979} and proposed an algorithm for computing deterministic Gittins indices to choose the arm to play. Though Gittins index  is proven to be optimal for discounted Bayesian bandits with Bernoulli rewards~\cite{gittins1979}, explicit computation of the indices is not always tractable and does not provide clear insights into what they look like and how they change as sampling proceeds~\cite{nino2011computing}. This motivated researchers to design computationally tractable algorithms~\cite{bubeck2012book} that still retain the asymptotic efficiency~\cite{lairobbins1985}. 

These algorithms can be classified into two categories: frequentist and Bayesian. Frequentist algorithms use the history obtained as the number of arm plays and corresponding rewards obtained to compute point estimates of the fitness index to choose an arm. UCB~\cite{auer2002finite}, UCB-tuned~\cite{auer2002finite}, KL-UCB~\cite{garivier2011klucb}, KL-UCB-Exp~\cite{garivier2011klucb}, KL-UCB$^+$~\cite{kaufmann2018} are examples of frequentist algorithms. These algorithms are designed by the philosophy of optimism in face of uncertainty. This methodology prescribes to act as if the empirically best choice is truly the best choice. Thus, all these algorithms overestimate the expected reward of the corresponding arms in form of frequentist indices.

Bayesian algorithms encode available information on the reward generation process in form of a prior distribution. For stochastic bandits, this prior consists of $K$ belief distributions on the arms. The history obtained by playing the bandit game is used to update the posterior distribution. This posterior distribution is further used to choose the arm to play. Thompson sampling~\cite{thompson1933}, information-directed sampling~\cite{russo2014information}, Bayes-UCB~\cite{kaufmann2018}, and BelMan are Bayesian algorithms.


In a variant of the stochastic bandit problem, called the \emph{pure
exploration bandit} problem~\cite{bubeck2009pure}, the goal of the
gambler is solely to accumulate information about the arms. 
In another variant of the stochastic bandit problem, the gambler interacts with the bandit in two consecutive phases of pure exploration and exploration--exploitation. \cite{putta2017twophaseepisodic} named this variant the \emph{two-phase reinforcement learning} problem. Two-phase reinforcement learning gives us a middle ground between
model-free and model-dependent approaches in decision making which is often the path taken by a practitioner~\cite{faheem2015adaptive}.
As frequentist methods are well-tuned for exploration-exploitation bandits, a different set of algorithms need to be developed for pure exploration bandits~\cite{bubeck2009pure}. \cite{kawale2015efficient} pointed out the lack of Bayesian methods to do so. This motivated recent developments of Bayesian algorithms~\cite{russo2016simple} which are modifications of their exploration--exploitation counterparts such as Thompson sampling. BelMan leverages its geometric insight to manage the pure exploration bandits only by turning the exposure to infinity. Thus, it provides a single framework to manage the pure exploration, exploration--exploitation, and two-phase reinforcement learning problems only by tuning the exposure.
\section{Conclusion}\label{sec:conc}
BelMan implements a generic Bayesian information-geometric approach for stochastic multi-armed bandit problems. It operates in a statistical manifold constructed by the joint distributions of beliefs and rewards. Their barycentre, the pseudobelief-reward, summaries the accumulated information and forms the basis of the exploration component. The algorithm is further extended by composing the pseudobelief-reward distribution with a reward distribution that gradually concentrates on higher rewards by means of a time-dependent function, the exposure. 
In short, BelMan addresses the issue of the adaptive balance of exploration–exploitation from the perspective of information representation, accumulation, and balanced induction of exploitative bias. Consequently, BelMan can be uniformly tuned to support pure exploration, exploration--exploitation, and two-phase reinforcement learning problems.
BelMan, when instantiated to rewards modelled by any distribution of the exponential family, conveniently leads to analytical forms that allow derivation of a well-defined and unique projection as well as to devise an effective and fast computation.
In queueing bandits, the agent tries and minimises the queue length while
also learning the unknown service rates of multiple servers. Comparative
performance evaluation shows BelMan to be more stable and efficient than
existing algorithms in the queueing bandit literature.

We are investigating the analytical asymptotic efficiency and stability of BelMan. We are also investigating how BelMan can be extended to other settings such as dependent arms, non-parametric distributions and continuous arms.
\section*{Software}
For the code of the queueing bandits, check: \href{https://github.com/Debabrota-Basu/QBelMan}{https://github.com/Debabrota-Basu/QBelMan}.
\section*{Acknowledgement}
We would like to thank Jonathan Scarlett for valuable discussions. This work is partially supported by WASP-NTU grant, the National University of Singapore Institute for Data Science project WATCHA, and Singapore Ministry of Education project Janus.
%
%
\bibliographystyle{alpha}
\bibliography{reference}  
\end{document}